\pdfoutput=1

\documentclass[11pt]{article}

\usepackage{ACL2023}

\usepackage{times}
\usepackage{latexsym}

\usepackage[T1]{fontenc}

\usepackage[utf8]{inputenc}

\usepackage{microtype}

\usepackage{inconsolata}

\usepackage{comment}
\usepackage{multicol}
\usepackage{multirow}
\usepackage{booktabs}
\usepackage{xspace}
\usepackage{cleveref}
\usepackage{tabularx}
\usepackage{graphicx}
\usepackage{enumitem}
\usepackage[export]{adjustbox}
\usepackage[htt]{hyphenat}
\usepackage{makecell}
\usepackage[normalem]{ulem}
\useunder{\uline}{\ul}{}
\usepackage{xurl}
\usepackage{subcaption}

\author{Chieh-Yang Huang,\textsuperscript{1*} Ting-Yao Hsu,\textsuperscript{1*} Ryan Rossi,\textsuperscript{2} Ani Nenkova,\textsuperscript{2} Sungchul Kim,\textsuperscript{2} \\
\textbf{Gromit Yeuk-Yin Chan,\textsuperscript{2} Eunyee Koh,\textsuperscript{2} Clyde Lee Giles,\textsuperscript{1} Ting-Hao `Kenneth' Huang\textsuperscript{1}} \\
  \textsuperscript{1}Pennsylvania State University, University Park, PA, USA. \\
  \texttt{\{chiehyang,txh357,clg20,txh710\}@psu.edu}\\
  \textsuperscript{2}Adobe Research, San Francisco, CA, USA. \\
  \texttt{\{ryrossi,nenkova,sukim,ychan,eunyee\}@adobe.com}
}

\begin{document}

%

\title{Summaries as Captions: Generating Figure Captions for Scientific Documents with Automated Text Summarization}


\definecolor{ao(english)}{rgb}{0.0, 0.5, 0.0}


\newcommand{\kenneth}[1]{}
\newcommand{\cy}[1]{}
\newcommand{\ed}[1]{}
\newcommand{\ryan}[1]{}
\newcommand{\gromit}[1]{}
\newcommand{\sukim}[1]{}

\newcommand{\oldDataset}{\textsc{SciCap}\xspace}

\newcommand{\system}{\textsc{LimitedInk}\xspace}

\newcommand{\dnns}{deep neural networks\xspace}
\newcommand{\Dnns}{Deep neural networks\xspace}
\newcommand{\dnn}{deep neural network\xspace}
\newcommand{\Dnn}{Deep neural network\xspace}
\newcommand{\DNN}{Deep Neural Network\xspace}
\newcommand{\DNNs}{Deep Neural Networks\xspace}

\newcommand{\mint}{model interpretation\xspace}
\newcommand{\mints}{model interpretations\xspace}
\newcommand{\Mint}{Model interpretation\xspace}
\newcommand{\Mints}{Model interpretations\xspace}

\newcommand{\inter}{interpretation\xspace}
\newcommand{\inters}{interpretations\xspace}
\newcommand{\Inter}{Interpretation\xspace}
\newcommand{\Inters}{Interpretations\xspace}

\newcommand{\eg}{{\it e.g.}}
\newcommand{\ie}{{\it i.e.}}
\newcommand{\etal}{{\it et al.}}

\newcommand{\sfigs}{scientific figures\xspace}
\newcommand{\sfig}{scientific figure\xspace}

\maketitle
\def\thefootnote{*}\footnotetext{Equal contribution.}\def\thefootnote{\arabic{footnote}}

\begin{abstract}
    Good figure captions help paper readers understand complex scientific figures.
Unfortunately, even published papers often have poorly written captions.
Automatic caption generation could aid paper writers by providing good starting captions that can be refined for better quality.
Prior work often treated figure caption generation as a vision-to-language task.
In this paper, we show that it can be more effectively tackled as a \textbf{text summarization} task in scientific documents.
We fine-tuned PEGASUS, a pre-trained abstractive summarization model, to specifically summarize figure-referencing paragraphs (\eg, ``Figure 3 shows...'') into figure captions.
Experiments on large-scale arXiv figures show that our method outperforms prior vision methods in both automatic and human evaluations.
We further conducted an in-depth investigation focused on two key challenges: {\em (i)} the common presence of low-quality author-written captions and {\em (ii)} the lack of clear standards for good captions.
Our code and data are available at: \url{https://github.com/Crowd-AI-Lab/Generating-Figure-Captions-as-a-Text-Summarization-Task}.

\end{abstract}

\section{Introduction\label{sec:introduction}}

\begin{figure*}[t]
    \centering
    \includegraphics[width=0.99\textwidth]{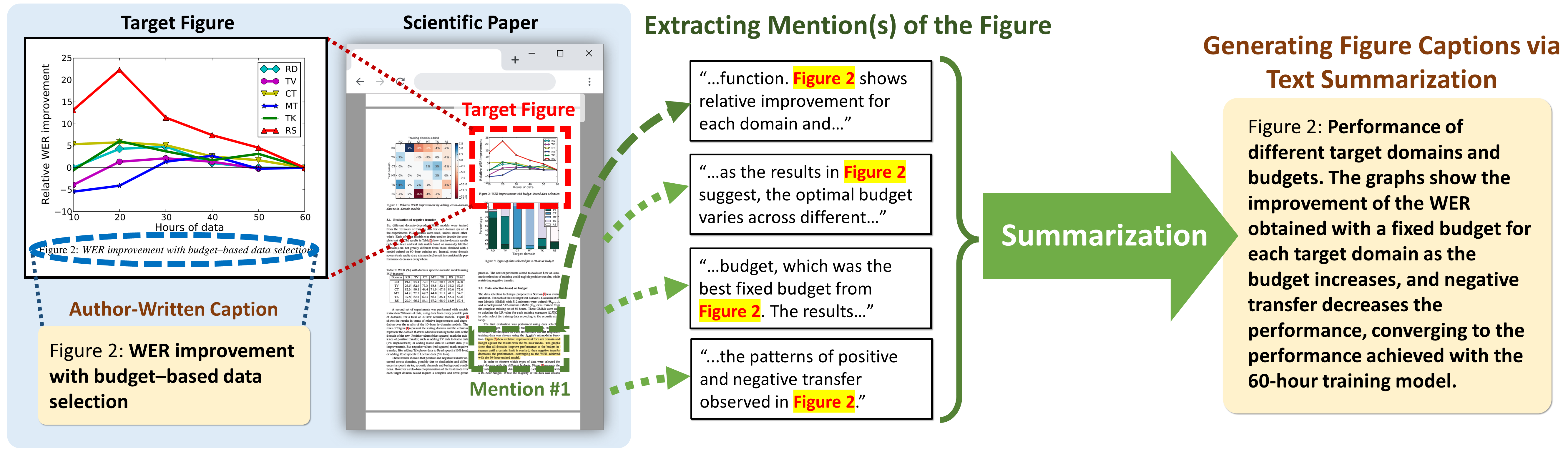}
    \vspace{-3mm}
    \caption{Figure captioning can be addressed as a text-summarization task. The figure’s caption can be generated by summarizing the paragraphs mentioning the figure. The caption is generated by the model Pegasus$_{P+O+B}$. The example shown in this figure is extracted from the paper~\cite{doulaty2015data}.}
   \label{fig:overview}
   \vspace{-2mm}
\end{figure*}

In scientific documents, effective figure captions help readers understand complex figures like bar charts, line charts, or pie charts.
These captions describe the images and often include necessary context from the document's full text~\cite{durbin2004effective}.
Unfortunately, even published papers often have poorly-written captions.
As per our analysis (\Cref{sec:quality-annotation}), around 53.88\% of line charts' captions in arXiv \texttt{cs.CL} papers are found to be unhelpful for NLP readers.
Automatic caption generation could aid paper writers by providing good starting captions that can be refined for better quality.

Previous research typically approached figure caption generation as a \textit{vision-to-language} task, \ie, creating captions based on the image.
For instance, \citet{hsu-etal-2021-scicap-generating} used an end-to-end approach with CNN+RNN structures, which extracted feature representation from the image and converted it into caption text. 
\citet{qian2021generating} took a slightly different approach: first understanding what is in the image, pulling out key information, and then using a preset template to create the caption.
However, although achieving some success in synthetic data~\cite{kahou2017figureqa,kafle2018dvqa,chen2019figure,zhu2021autochart}, these approaches often struggled to caption real-world figures.
For example, \citet{hsu-etal-2021-scicap-generating}'s end-to-end approach, trained and tested using arXiv figures, achieved a BLEU-4 score of only 2.91.

In this paper, we argue that figure captioning in scientific documents can be more effectively tackled as a \textbf{text-summarization task}: 
The caption can be generated by \textbf{summarizing the paragraphs mentioning the figure} (as shown in Figure~\ref{fig:overview}.)
Scientific figures typically come with extensive text in the scientific document that can aid caption generation.
Our analysis (Section~\ref{sec:motivating-analysis}) shows that, in arXiv, over 75\% of words in figure captions can be aligned with the words in the paragraphs referencing those figures, which motivates our approach.
The automatic evaluation shows that summarizing paragraphs referencing the figures results in better captions than prior vision-based methods.
In a human evaluation by external domain experts, our best-performing model's captions were preferred over the original captions 46.67\% of the time.

We further conducted an in-depth investigation focused on two key challenges: {\em (i)} the common presence of low-quality author-written captions and {\em (ii)} the lack of clear standards for good captions.
Surprisingly, 53.88\% of the author-written captions in our sample was deemed unhelpful.
This has implications for the design of future captioning systems, underscoring the influence of data quality on captioning performance.

\section{Related Work\label{sec:related-work}}




Prior figure captioning works can be broadly categorized into two approaches:
caption generation {\em (i)} based on the image of the figure or {\em (ii)} based on the data chart underlying the figure.

Earlier image-based approaches focused on automated image understanding, which involved parsing images to extract the figure's key attributes and converting parsed data into captions, \eg, using predefined templates~\cite{kahou2017figureqa,kafle2018dvqa,Methani_2020_WACV,qian2021generating,siegel2016figureseer}.
Recently, with the advance of deep learning, more works are adopting an end-to-end paradigm, generating captions straight from the neural representations of images~\cite{9973197,ImageCLEFmedConceptOverview2021,hsu-etal-2021-scicap-generating,10.1145/3341162.3345601,kanthara2022chart,chen2019figure}.
Our work contrasts with prior studies by focusing on text to generate captions instead of visuals. 
To the best of our knowledge, no existing figure-caption datasets explicitly contain the figures' accompanying documents~\cite{ImageCLEFmedConceptOverview2021,hsu-etal-2021-scicap-generating,10.1145/3341162.3345601}, as this task has generally been approached as a vision task.
Most recently, a knowledge-augmented image captioning method that uses both image and text data was introduced~\cite{scicap-plus}, suggesting the potential of using text from documents.

Some approaches generate captions using the underlying tabular data of a figure rather than the figure's image. 
Earlier approaches often employed rule-based techniques~\cite{corio1999generation, demir2008generating, fasciano1996postgraphe, mittal1998describing}, while newer ones favor learning-based methods~\cite{barzilay-lapata-2005-collective,wiseman-etal-2017-challenges,moraes2014generating, zhu2021autochart, kanthara2022chart,obeid-hoque-2020-chart,reiter2005choosing,parikh-etal-2020-totto,chen-etal-2020-logical,gong-etal-2019-enhanced,su-etal-2021-plan-generate,chen-etal-2020-logic2text}. 
Despite these approaches' ability to utilize tabular and meta data, they necessitate access to the figure's raw data. 
Contrarily, our work uses the rich textual information in scientific documents to generate captions.

\section{Problem Statement and Terminology\label{sec:problem-statement}}

A document $D$ contains $n$ figures, $F_1$ to $F_n$, where $F_i$ has a caption $C_i$ that was written by the document author.
In document $D$, $j$ sentences, $M_{i,1}$ to $M_{i,j}$, explicitly mention $F_i$ (\eg, ``As shown in $F_i$...'').
The objective of this work is to automatically generate a high-quality caption, $C_i'$, for figure $F_i$ using only its mentions ($M_{i,1}$ to $M_{i,j}$) and the surrounding text of the mentions in document $D$.

In the rest of the paper, we use these terms:

\begin{table*}[t]
    \centering \small
    \addtolength{\tabcolsep}{-1.1mm}
    \begin{tabular}{@{}l@{\kern5pt}l@{\kern3pt}ccccccccccccccc@{}}
    \toprule
    \multicolumn{2}{@{}l}{\multirow{2}{*}{\textbf{Source}}} & \multicolumn{2}{c}{\textbf{Random}} & \multicolumn{2}{c}{\textbf{Mention}} & \multicolumn{2}{c}{\textbf{Paragraph}} & \textbf{OCR} & \multicolumn{2}{c}{\textbf{Window{[}0, 1{]}}} & \multicolumn{2}{c}{\textbf{Window{[}1, 0{]}}} & \multicolumn{2}{c}{\textbf{Window{[}1, 1{]}}} & \multicolumn{2}{c}{\textbf{Window{[}2, 2{]}}} \\ \cmidrule{3-17}
    \multicolumn{2}{c}{} & \textbf{S} & \textbf{P} & \textbf{-} & \textbf{+OCR} & \textbf{-} & \textbf{+OCR} & \textbf{-} & \textbf{-} & \textbf{+OCR} & \textbf{-} & \textbf{+OCR} & \textbf{-} & \textbf{+OCR} & \textbf{-} & \textbf{+OCR} \\ \midrule
    
    \multicolumn{2}{@{}l}{\textbf{Caption}} & 35.23 & 44.43 & 53.43 & 60.16 & 75.19 & \textbf{76.68} & 34.75 & 60.85 & 65.43 & 59.09 & 64.19 & 65.20 & 68.73 & 69.09 & 71.77 \\
    \multicolumn{2}{@{}l}{\textbf{Source}} & 32.52 & 19.52 & \textbf{39.51} & 18.78 & 12.53 & 9.39 & 20.79 & 30.49 & 17.19 & 32.40 & 17.33 & 25.10 & 15.55 & 19.84 & 13.45 \\ 
    \bottomrule
    \end{tabular}
    \addtolength{\tabcolsep}{+1.1mm}
    \caption{Macro coverage rates (percentage) between captions and relevant texts (S: Sentence and P: Paragraph). 
    Caption coverage gives the percentage of words in the caption that can be found in the source texts and vice versa (punctuation and stop words are excluded.)
    The results show that 76.68\% of the words in captions could be found in Paragraph+OCR,
    motivating us to generate captions by text summarization.
    }
    \label{tab:alignment-result}
    \vspace{-2mm}
\end{table*}

\begin{itemize} [leftmargin=*,itemsep=0mm]

    \item 
    A \textbf{``Mention''} refers to a sentence in a document that explicitly mentions the target figure, \eg, ``As shown in Figure 6...''
    If there are multiple Mentions, the first Mention is referred to. 


    \item 
    
    A \textbf{``Paragraph''} refers to a section of text containing a Mention. In this work, the boundaries of a Paragraph are determined by the \texttt{<p>} tag produced by PDF parsing.


    \item 
    Sentences near a Mention may contain relevant information, so we extracted $n$ preceding sentences and $m$ following sentences to form the \textbf{``Window[n,~m]''} text snippet.
    For instance, ``Window[1,~2]'' refers to a snippet of four sentences, including one preceding sentence, the Mention sentence, and two following sentences. 


    \item 
    An \textbf{``OCR''} refers to the textual information (\eg, legends, labels, etc.) extracted from the image, by optical character recognition (OCR) software.

\end{itemize}

\section{Dataset\label{sec:data}}

\kenneth{Add a few sentences to say (1) the remaining of the paper was built using SciCap, including motivating analysis and testing of the proposed methods and in-depth analysis. (2) We additionally add mention/paragraph and resplit the data (suggesting we did a lot more than just using it.)}

\kenneth{I agree with Gromit that it reads a bit sudden. We can add a sentence to explain that we want to introduce dataset here as it was used in Motivating Anlaysis too.}
\cy{done}


Before diving into our experiments and analyses, we first describe the dataset upon which our study is grounded.
Our results are based on a scientific figure caption dataset, \oldDataset, and several pre-processing steps to fit it into our workflow.

%
%
\oldDataset is a dataset that contains over 416,000 line charts and captions extracted from more than 290,000 arXiv papers~\cite{hsu-etal-2021-scicap-generating}. 
It was one of the first large-scale figure-captioning datasets based on real-world \sfigs. 
However, it does not contain the paragraphs that mention the figure.
To address this, we downloaded all the PDF files of the original arXiv papers used in \oldDataset and
extracted all the Mentions and Paragraphs as outlined in \Cref{sec:extract-mention}.
Detailed information on preprocessing, including the dataset resplit and OCR extraction, are described in \Cref{sec:appendix-data-preprocessing}.

\section{Motivating Analysis\label{sec:motivating-analysis}}




To understand the correlation between mentions and captions, we performed a series of analyses using the data described in \Cref{sec:data}.
Specifically, we investigated the extent to which the words in the figure captions are represented in the corresponding figure-mentioning paragraphs.
We used awesome-align~\cite{dou2021word} to obtain the alignment between the source texts 
(mentions, paragraphs, and OCRs) and captions.
Awesome-align compared the similarity of the words' contextual embeddings 
and assigned an alignment between words if the similarity passed a threshold. 
We used SciBERT~\cite{beltagy-etal-2019-scibert}
to obtain contextual embeddings and softmax threshold = 0.99 to reduce false alignments.

After obtaining the alignments, we computed what percentage of information in the caption could be found in the source texts.
The results shown in \Cref{tab:alignment-result} indicate that \textbf{76.68\% of the caption's information could be found in Paragraph and OCR,} motivating us to generate figure captions by summarizing Paragraph.
We also observed that a randomly selected sentence and paragraph from the same paper can cover 35.23\% and 44.43\% of the caption,
respectively, showing that there was some generic information-sharing across the paper.
We also conducted a study using the exact overlapping (\ie, BLEU score) in \Cref{sec:token-overlap-study}.

\section{Generating Figure Captions as a Text Summarization Task\label{sec:method}}

Figure~\ref{fig:overview} overviews the proposed pipeline. 
This section describes each step of the pipeline.

\subsection{Extracting Mentions and Paragraphs\label{sec:extract-mention}}
The system first extracts Mentions and their associated Paragraphs (as defined in Section~\ref{sec:problem-statement}.)
In this paper, we used Grobid~\cite{grobid}, a publicly-available tool for converting PDF files into structured XML documents, to extract plain text from the paragraphs (including the \texttt{<p>} tags) in each paper.
This plain text was then segmented into sentences using BlingFire~\cite{blingfire}.
We developed regular expressions to identify sentences mentioning specific figures. 
For instance, sentences such as ``As shown in Figure 6, ...'' were first identified and then linked to Figure 6. 
To assess the performance of these regular expressions, we conducted a manual evaluation of 300 samples from our experimental dataset.
The results showed a high level of precision (99.58\%) and recall (94.44\%).





\subsection{Generating Captions Using Text Summarization Models}
As shown in Figure~\ref{fig:overview}, our system then automatically summarizes all the extracted Mentions (or Paragraphs) into a figure caption.
In this work, we used PEGASUS, an abstractive summarization model~\cite{zhang2020pegasus}, and fine-tuned it on our dataset.
Five Pegasus models, Pegasus$_{M}$, Pegasus$_{P}$, Pegasus$_{O}$, Pegasus$_{M+O}$, and Pegasus$_{P+O}$, were trained utilizing five distinct input combinations, including
\textbf{
{\em (i)} Mention, 
{\em (ii)} Paragraph, 
{\em (iii)} OCR output of the target figure image, 
{\em (iv)} Mention+OCR, and 
{\em (v)} Paragraph+OCR.}
Pegasus$_{P+O}$ encompasses the most of relevant information in the document and thus is expected to yield the optimal summary.

Additionally, we built Pegasus$_{P+O+B}$, a specialized version of the model designed to be trained on a subset of higher-quality captions, \textbf{{\em (vi)} Paragraph+OCR-Better}.
Given the absence of reliable automated ways to assess the quality of captions, we followed a guideline from previous studies indicating that longer captions enhance reader comprehension~\cite{hartley2003single,gelman2002let}.
We trained the model using captions with 30 or more tokens.
The average caption length was 26.8 tokens, so we set 30 tokens as the threshold.
The training was performed using Paragraph+OCR inputs.




We identified two major challenges in generating captions for scientific figures in real-world scenarios.
We discuss these challenges in the following subsections, with an in-depth analysis in Section~\ref{sec:analysis}.


\subsubsection{Challenge 1: Addressing Unreliable Quality of Real-World Data}
Low-quality captions often occur in scholarly articles.
Our analysis (see \Cref{sec:analysis-challenge-1}) showed that 50\% of line charts' author-written captions in arXiv {\tt cs.CL} papers were deemed unhelpful by domain experts.
The impact of this unreliable data quality is that developers could train and test captioning models with unhelpful captions.
The lack of automatic methods for evaluating caption quality makes it hard to identify suitable training examples and eliminate poor ones.
To address this issue, 
we included Pegasus$_{P+O+B}$ that was trained on longer captions, which is suggested by literature to be more helpful to readers~\cite{hartley2003single,gelman2002let}.
To account for low-quality test data, we conducted both human and automatic evaluations.
The data quality of figure captions was analyzed and is presented in \Cref{sec:quality-annotation}.

\subsubsection{Challenge 2: Defining a Clear Standard for ``Good'' Figure Captions}
The deeper issue is the lack of a set of well-defined and actionable criteria for determining the usefulness of a figure caption.
Although there are guidelines for writing effective scientific figure captions~\cite{10.1371/journal.pcbi.1003833,doi:10.1021/acsenergylett.9b00253}, their translation into algorithmic models can be challenging.
From a modeling standpoint, the lack of a clear goal presents a challenge, as it is uncertain what to optimize for once fluency has been achieved.
In this paper, we focus on demonstrating the feasibility of generating captions via text summarization.
Although we did not incorporate specialized goals in the model, we examine the criteria for a ``good'' caption in Section~\ref{sec:quality-annotation}.

\section{Experimental Results\label{sec:result}}

\begin{table*}[]
\centering \small
\addtolength{\tabcolsep}{-1.0mm}
\begin{tabular}{@{}llccccccccccc@{}}
\toprule
\multirow{2}{*}{\textbf{Model}} & \multirow{2}{*}{\textbf{Feature}} & \multirow{2}{*}{\textbf{Length}} & \multicolumn{2}{c}{\textbf{Rouge-1 (F1)}} & \multicolumn{2}{c}{\textbf{Rouge-2 (F1)}} & \multicolumn{2}{c}{\textbf{Rouge-L (F1)}} & \multicolumn{2}{c}{\textbf{MoverScore}} & \multicolumn{2}{c}{\textbf{BERTScore}} \\ \cmidrule(l){4-13}
 &  &  & \textbf{Score} & \textbf{Norm} & \textbf{Score} & \textbf{Norm} & \textbf{Score} & \textbf{Norm} & \textbf{Score} & \textbf{Norm} & \textbf{Score} & \textbf{Norm} \\ \midrule
\multirow{6}{*}{\textbf{Reuse}} & \textbf{M} & 33.2 & .291 & 1.346 & .139 & 1.790 & .239 & 1.401 & .535 & 1.023 & .628 & 1.064 \\
 & \textbf{P} & 238.3 & .171 & 1.042 & .089 & 1.006 & .134 & 1.030 & .503 & 1.004 & .567 & 1.008 \\
 & \textbf{W{[}0, 1{]}} & 50.3 & .281 & 1.216 & .132 & 1.509 & .224 & 1.273 & .529 & 1.016 & .620 & 1.048 \\
 & \textbf{W{[}0, 2{]}} & 68.0 & .259 & 1.129 & .123 & 1.341 & .205 & 1.186 & .524 & 1.013 & .611 & 1.034 \\
 & \textbf{W{[}1, 1{]}} & 67.8 & .266 & 1.156 & .124 & 1.346 & .204 & 1.183 & .524 & 1.012 & .613 & 1.037 \\
 & \textbf{W{[}2, 2{]}} & 98.7 & .235 & 1.082 & .112 & 1.179 & .180 & 1.105 & .517 & 1.007 & .600 & 1.020 \\ \midrule
\multirow{6}{*}{\textbf{Pegasus}} & \textbf{M} & 12.2 & .321 & 1.898 & .153 & 2.907 & .283 & 1.971 & .553 & 1.065 & .654 & 1.158 \\
 & \textbf{M+O} & 12.8 & .331 & 1.909 & .161 & 2.945 & .292 & 1.993 & .556 & 1.071 & .661 & 1.166 \\
 & \textbf{P} & 14.0 & {\ul .374} & {\ul 2.067} & {\ul .205} & {\ul 3.507} & {\ul .334} & {\ul 2.201} & {\ul .570} & {\ul 1.095} & {\ul .682} & {\ul 1.196} \\
 & \textbf{P+O} & 14.0 & \textbf{.381} & \textbf{2.106} & \textbf{.212} & \textbf{3.635} & \textbf{.340} & \textbf{2.242} & \textbf{.571} & \textbf{1.097} & \textbf{.685} & \textbf{1.202} \\
 & \textbf{P+O+B} & 38.3 & .321 & 1.452 & .154 & 1.916 & .265 & 1.537 & .546 & 1.044 & .639 & 1.082 \\
 & \textbf{O} & 12.1 & .133 & 0.789 & .026 & 0.495 & .119 & 0.828 & .518 & 0.998 & .561 & 0.993 \\ \midrule
\textbf{TrOCR} & \multirow{2}{*}{\textbf{Figure}} & 10.0 & .220 & 1.464 & .073 & 1.653 & .195 & 1.502 & .534 & 1.033 & .610 & 1.096 \\
\textbf{BEiT+GPT2} &  & 15.8 & .164 & 0.864 & .042 & 0.666 & .144 & 0.917 & .529 & 1.013 & .592 & 1.031 \\
\bottomrule
\end{tabular}
\addtolength{\tabcolsep}{+1.0mm}
\caption{Task Performance with the \textbf{best} and {\ul second-best} results highlighted. Pegasus$_{P+O}$, the text-summarization model with all available information (Paragraph+OCR), performed the best in all four metrics. Pegasus$_{P+O+B}$, the model trained with better captions, however, got lower scores.}
\label{tab:result-summarization}
\vspace{-3mm}
\end{table*}

\paragraph{A Simple Baseline: Using Extracted Mentions as Captions.}
Motivated by our information overlap study (\Cref{sec:motivating-analysis}),
we created the \textbf{Reuse} baselines. 
These baselines simply repurpose portions of the input text as the prediction.


\paragraph{Vision-to-Language Baselines.}
The vision-to-language generation treated this task as an image-captioning task 
that took the scientific figure image as input and generated a text to describe it.
We compared two vision-to-language models as baselines.
First, we built a sequence-to-sequence model by combining BEiT~\cite{bao2022beit} and GPT-2~\cite{radford2019language}.
We also selected the TrOCR~\cite{li2021trocr} model, a transformer-based sequence-to-sequence model pre-trained for OCR tasks.
Compared to image encoders like ViT~\cite{dosovitskiy2020vit} and BEiT~\cite{bao2022beit}, which were trained on photos, OCR models trained on printed and handwritten documents align more closely with the scientific paper domain.
All figures from \oldDataset (106,391 training samples) were used for training since no mentions were required.



\begin{figure*}[t]
    \centering
    \includegraphics[width=0.48\linewidth]{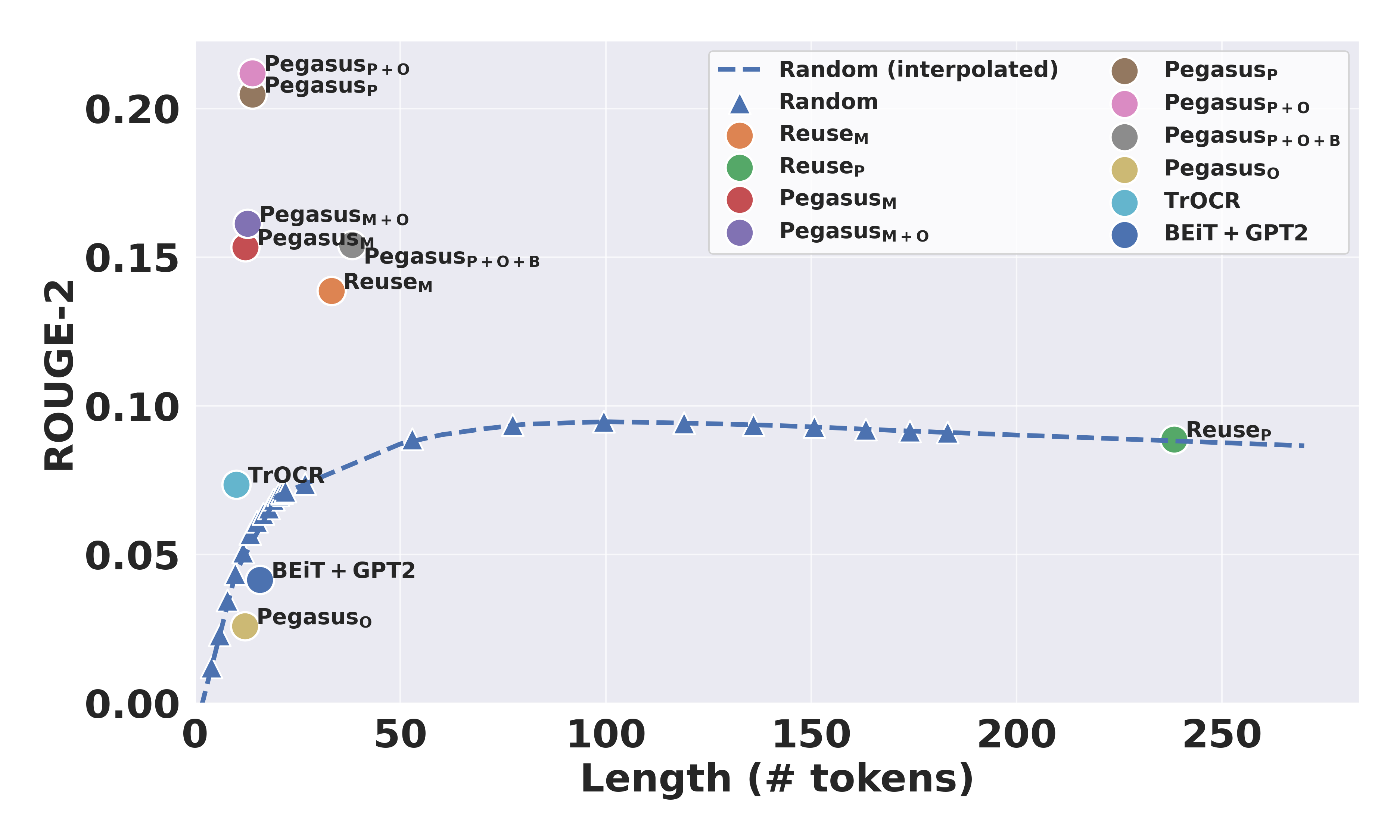}
    \includegraphics[width=0.48\linewidth]{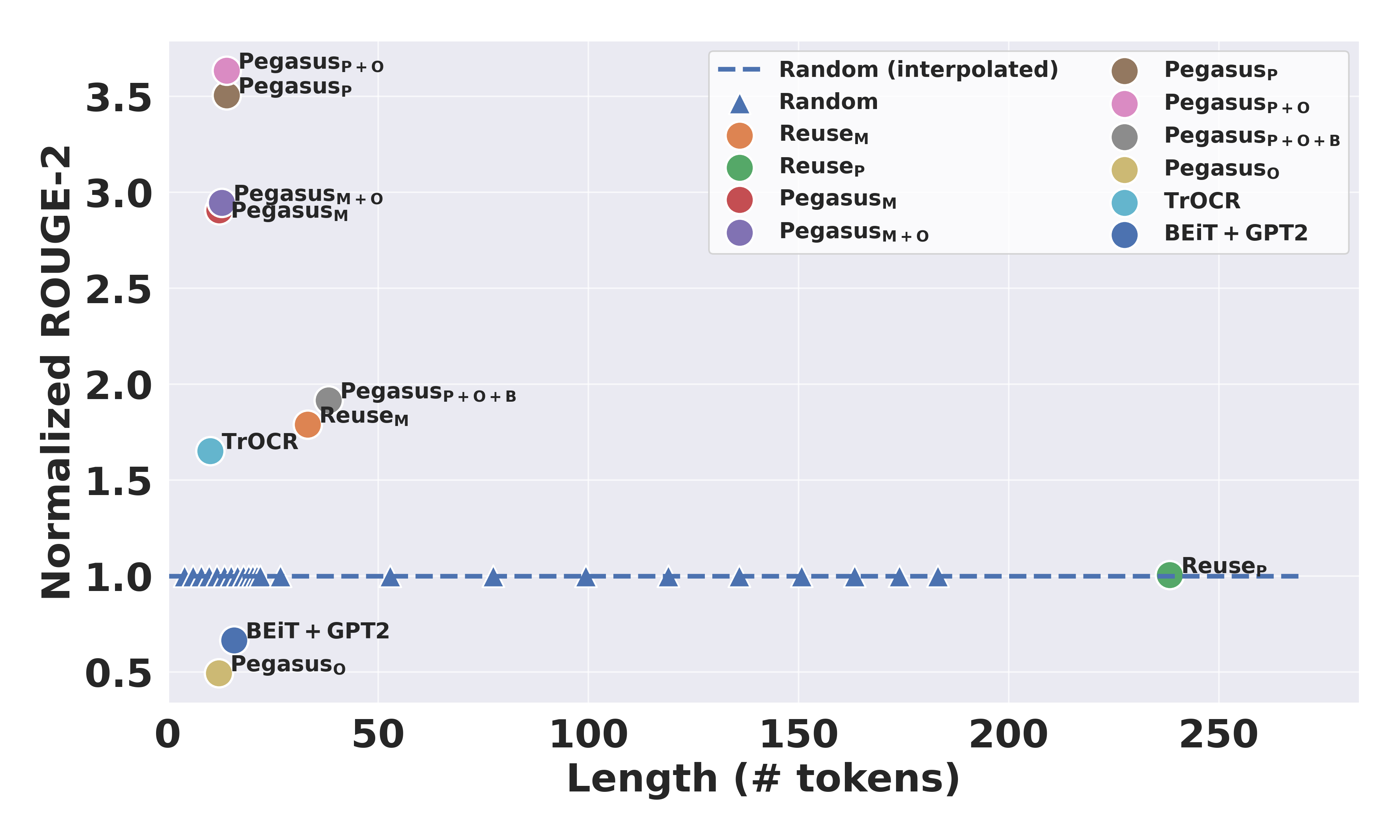}
    \vspace{-3mm}
    \caption{The relationship between average text length and ROUGE-2 score (left: original ROUGE-2; right: normalized ROUGE-2). The random baseline in the left figure shows that text lengths and scores are not independent. For example, when the predicted text is shorter than 50 tokens, predicting longer texts generally results in a higher ROUGE-2 score. The normalized scores indicate the proposed system's performance gain over the random baseline of the same length. Pegasus$_{P+O+B}$ and Reuse$_{M}$ get closer to TrOCR after normalization, suggesting the need for normalization for accurate interpretation of results.} 
    \label{fig:length-performance}
    \vspace{-3mm}
\end{figure*}

\paragraph{Experimental Setup.}
\label{sec:training-setup}

A total of 14 methods were included for comparison: 
six reuse baselines with six input variations
(M, P, W[0, 1], W[0, 2], W[1, 1], and W[2, 2]);
five text summarization models with five inputs
(M, M+O, P, P+O, and O);
one text summarization model using P+O with controlled data quality; 
and two vision-to-language models (BEiT+GPT-2 and TrOCR). 
Note that we use subscripts of M, P, W, O, B to denote the input features: 
Mention, Paragraph, Window, OCR, and Better data quality, respectively.
The model training details and decoding configuration are provided in \Cref{sec:appendix-model-training-details}.




\subsection{Automatic Evaluation Results}
\paragraph{Conventional Automatic Evaluation.}
We used F1 of ROUGE-1, ROUGE-2, ROUGE-L~\cite{lin-2004-rouge,nallapati-etal-2016-abstractive}, 
MoverScore~\cite{zhao-etal-2019-moverscore}, and BERTScore for automatic evaluation. 
When computing ROUGE scores using rouge-score~\cite{rouge-scorer},
we turned all text into lower case and stem words.
As both MoverScore and BERTScore are based on the semantic similarity,
we obtained contextual embeddings from SciBERT~\cite{beltagy-etal-2019-scibert}.



\paragraph{Automatic Evaluation with Normalization Over Caption Length.}
ROUGE F1 tends to favor longer texts within a certain length, leading to a skewed comparison where models generating longer texts receive higher scores~\cite{sun-etal-2019-compare}.
We followed \citet{sun-etal-2019-compare}'s approach of normalizing the scores 
with the corresponding random baseline that generates texts of the same length. 
\begin{equation}
    \small
    Score_{normalized} = \frac{Score}{Random(length)}
\end{equation}
where $length$ is the average length of the texts generated by the target system.
We estimated $Random(length)$ by applying linear interpolation on several (length, random score) pairs.
The (length, random score) pairs were obtained by randomly selecting a certain number of sentences (1, 2, ..., 10 sentences) from the input paragraph as the prediction. 
To get random scores of texts shorter than a single sentence (around 30 tokens), 
we truncated sentences to the desired length (4, 6, ..., 30 tokens). 
For each length setting, we ran 10 different random seeds and reported the average. 
The Random line in \Cref{fig:length-performance}-Left shows the behavior of ROUGE-2
favoring longer texts within 50 tokens.\footnote{Similar trends for ROUGE-1, ROUGE-L, MoverScore, and BERTScore 
are included in \Cref{sec:additional-results}.}
The normalized scores, as shown in \Cref{fig:length-performance}-Right, 
clearly indicate the superiority of our proposed model over the random baseline.

\Cref{tab:result-summarization} shows the normalized automatic evaluation results. 
Overall, \textbf{Pegasus$_{P+O}$, the text-summarization model with all available information (Paragraph+OCR), 
achieved the best performance in all three metrics.}
Pegasus$_{P+O+B}$, the model using the same information but trained on a better subset of captions (Paragraph+OCR-Better), did not perform well. 
We hypothesized this was due to half of the test data comprising poor captions (refer to \Cref{sec:quality-annotation}). 
This was validated by examining performance shifts in different quality beams (\Cref{sec:auto-eval-quality-beam}) and conducting a human evaluation (\Cref{sec:human-eval}).
Meanwhile, Reuse$_{M}$, the Reuse baseline with Mention, outperforms other Reuse baselines.
Its performance declined as context sizes grew and shifted.

%


\subsection{Human Evaluation Results\label{sec:human-eval}}


\paragraph{Pilot MTurk Study to Select Top Models.}
\cy{Probably emphasize why we choose these four systems for comparison}
Before the main human evaluation, we ran a pilot study on Amazon Mechanical Turk (MTurk) to identify any apparently underperforming baselines for exclusion in the final study, simplifying the main human evaluations.
In this study, we asked MTurk workers to carefully read a figure and select the \textit{worst} figure caption among 
{\em (i)} TrOCR, 
{\em (ii)} Pegasus$_{P+O}$, 
{\em (iii)} Pegasus$_{P+O+B}$, and 
{\em (iv)} ground-truth caption. 
Ninety figures without errors were randomly sampled from our annotated set (\ie, figures from \texttt{cs.CL} arXiv papers in \Cref{sec:quality-annotation}) for the study.
For each of the figures, we recruited 20 MTurk workers to judge.\footnote{Four MTurk qualifications were used: Locale (US Only), HIT Approval Rate ($\geq$98\%), Number of Approved HITs ($\geq$3000), and the Adult Content Qualification. The payment for each task was set to 0.09 (hourly wage = \$10 dollars).}
We report the number of majority votes (when tied, we counted all captions with the highest votes as the worst) and the average number of votes in \Cref{tab:mturk-eval}.
Results indicated that TrOCR's caption won the majority vote 41 out of 90 times, with its average vote count significantly exceeding others.
Hence, we excluded TrOCR from our formal human evaluation.

\paragraph{Main Human Evaluation with Domain Experts.}
Three Ph.D. students with NLP backgrounds (who are not coauthors) were recruited as human judges, 
as it is hard for those without basic domain understanding to evaluate captions.
This study has been approved by the IRB office of the authors' institute.
The same 90 figures used in the pilot MTurk study were used again.
We asked the human judges to compare each figure's 
(\textit{i}) Pegasus$_{P+O}$, (\textit{ii}) Pegasus$_{P+O+B}$, and (\textit{iii}) ground-truth caption. 
The judges were asked to rank the captions based on how strongly they agreed with this statement:
``When I read the paper, this caption can help me understand the message that the figure tries to convey.'' 
\Cref{fig:ui-compare} (see \Cref{sec:app-interface}) shows the interface the human judges used. 

\begin{table}[]
\centering \small
\begin{tabular}{@{}l@{}c@{\kern2pt}c@{\kern2pt}c@{\kern2pt}c@{\kern2pt}c@{}}
\toprule
 \multirow{2}{*}{\textbf{n = 90}} & \multirow{2}{*}{\makecell{\textbf{\#Maj.} \\ \textbf{Votes}$\downarrow$}} & \multirow{2}{*}{\makecell{\textbf{Avg.} \\ \textbf{Votes}$\downarrow$}} & \multicolumn{3}{c}{\textbf{T-Test over Avg. Votes}} \\ \cmidrule{4-6}
 &  &  & \textbf{Peg}$_{P+O}$ & \textbf{Peg}$_{P+O+B}$ & \textbf{Caption} \\ \midrule
\textbf{TrOCR} & 41 & 5.99 & \textless .001*** & .006** & .001** \\
\textbf{Peg}$_{P+O}$ & 20 & 4.54 & - & .253 & .973 \\
\textbf{Peg}$_{P+O+B}$ & 24 & 4.93 & - & - & .318 \\
\textbf{Caption} & 19 & 4.53 & - & - & - \\
\bottomrule
\end{tabular}
\caption{The result of the pilot Mturk study. When tied, all captions with the highest votes were counted as the worst for \#Majority votes. TrOCR is significantly worse than other approaches when rated by crowd workers.}
\label{tab:mturk-eval}
\vspace{-3mm}
\end{table}

\begin{table}[]
    \centering \small
    
    \begin{tabular}{@{}lccc@{}}
    \toprule
    \multirow{2}{*}{\textbf{n = 90}} & \multirow{2}{*}{\makecell{\textbf{Avg.} \\ \textbf{Ranking}$\downarrow$}} & \multicolumn{2}{c}{\textbf{T-Test on Avg. Ranking}} \\ \cmidrule{3-4}
     &  & \textbf{Peg}$_{P+O+B}$ & \textbf{Caption} \\ \midrule
    \textbf{Peg}$_{P+O}$ & 2.152 & .016* & .015* \\
    \textbf{Peg}$_{P+O+B}$ & 1.930 & - & .923 \\
    \textbf{Caption} & \textbf{1.919} & - & - \\ \bottomrule
    \end{tabular}
    
    \caption{Average ranking of the human evaluation. Pegasus$_{P+O+B}$ was rated significantly better than Pegasus$_{P+O}$ and was at the same level as the ground-truth caption.}
    \vspace{-2mm}
    \label{tab:human-eval}
    \vspace{-4mm}
\end{table}

\Cref{tab:human-eval} shows the results of average ranking (from 1 to 3). 
Overall, \textbf{the ground-truth caption and Pegasus$_{P+O+B}$ were ranked similarly} (1.919 vs. 1.930 with p-value = 0.923). 
\textbf{Humans also favored Pegasus$_{P+O+B}$ over Pegasus$_{P+O}$ significantly} (1.919 vs. 2.152 with p-value = 0.016).
This supports our heuristic for automatically determining caption quality based on length 
and aligns with previous findings that longer captions improve reader comprehension~\cite{hartley2003single,gelman2002let}. 
However, we found that the task of caption ranking poses a challenge, as evidenced by the lower correlations between raters, with Kendall's tau values of 0.133, 0.148, and 0.274, and Spearman's rho values of 0.128, 0.156, and 0.317.
This highlights the complexity of the task and suggests that scaling human evaluation across domains might be difficult.
Different preferences over captions, such as length, could lead to lower agreement among raters.

\section{In-Depth Analysis\label{sec:analysis}}

We conducted an in-depth investigation focused on two key challenges: {\em (i)} the common presence of low-quality author-written captions and {\em (ii)} the lack of clear standards for good captions.

\paragraph{Quality Annotation Procedure.}
We manually annotated 438 captions in the Computation and Language domain (\texttt{cs.CL}) from the test set. 
\Cref{fig:ui-rating} (see \Cref{sec:app-interface}) shows the interface we used, in which the title, abstract, 
and PDF file of the paper were shown alongside the target figure's image, caption, and questions.
For each caption, we asked the annotators (coauthors) to rate four aspects using a five-point Likert scale:


\begin{table}[t]
    \small
    \centering
    \addtolength{\tabcolsep}{-1mm}
    
    \begin{tabular}{@{}lcccc@{}}
    \toprule
    \multicolumn{1}{c}{\textbf{}} & \textbf{Agree} & \textbf{Disagree} & \textbf{Total} & \textbf{Agree Percentage} \\ \midrule
    \textbf{Helpfulness}        & 184 & 215 & 399 & 46.12\% \\
    \textbf{Image-Text}         & 338 & 61 & 399 & 84.71\% \\
    \textbf{Visual-Desc}        & 64 & 335 & 399 & 16.04\% \\
    \textbf{Takeaway}           & 74 & 325 & 399 & 18.55\% \\ \bottomrule
    \end{tabular}
    \addtolength{\tabcolsep}{+1mm}
    \vspace{-1mm}

    \caption{Results of the manual annotation.
    More than 50\% of the captions were annotated as unhelpful. (Out of the initial 438 figure captions, we excluded those with extraction or classification errors, \eg, incomplete images, leaving us with only 399 captions.)}
    \label{tab:quality-annotation-stat}
    \vspace{-2mm}
    
\end{table}

\vspace{-.3pc}

\begin{itemize}  [leftmargin=*,itemsep=0mm] 
    \item \textbf{Image-Text.} The caption included named entities or important words/numbers in the figure (\eg, title, legends, labels, etc.).
    \vspace{-.3pc}

    \item \textbf{Visual-Description.} The caption included some visual characteristics of the figure (\eg, color, shape, trend, etc.).
    \vspace{-.3pc}

    \item \textbf{Takeaway.} The caption explicitly stated the high-level takeaway message or the conclusion that the figure attempted to convey.
    \vspace{-.3pc}

    \item \textbf{Helpfulness.} ``The caption helped me understand the message that the figure attempted to convey''.
\end{itemize}
\vspace{-.3pc}
%
The annotated data was consolidated by grouping ``Strongly Agree'' and ``Agree'' as ``\texttt{[Agree]}'' and grouping ``Neutral'', ``Disagree'', and ``Strongly Disagree'' as ``\texttt{[Disagree]}''.
The results of this consolidation are presented in Table~\ref{tab:quality-annotation-stat}.

\begin{figure*}[t]
    \centering
    \includegraphics[width=0.49\linewidth]{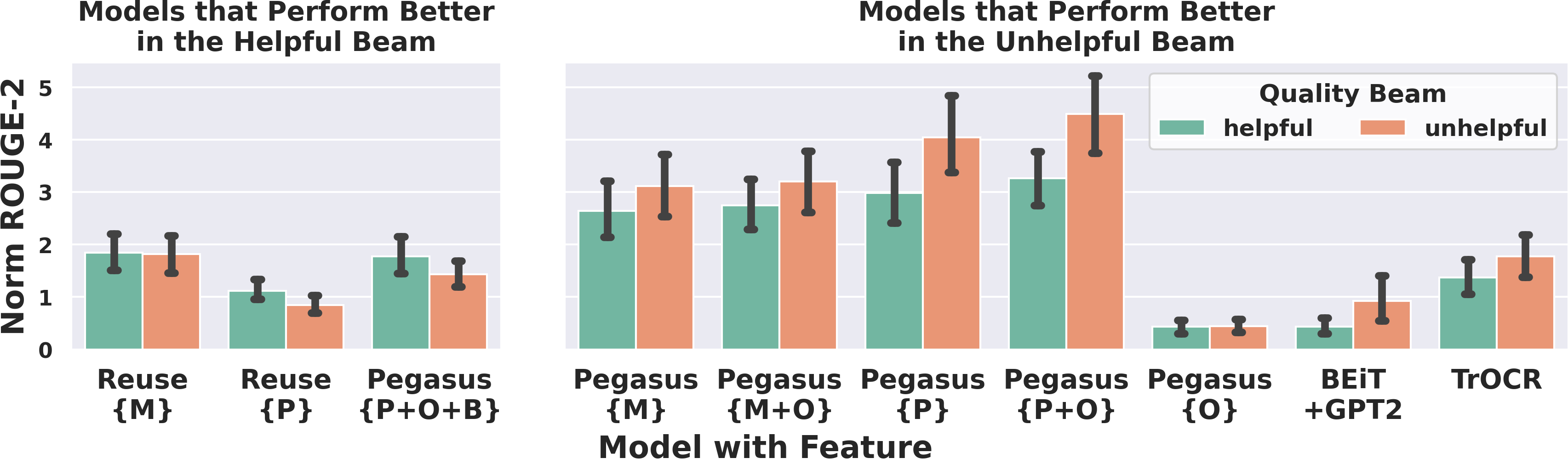}
    \hfill
    \includegraphics[width=0.49\linewidth]{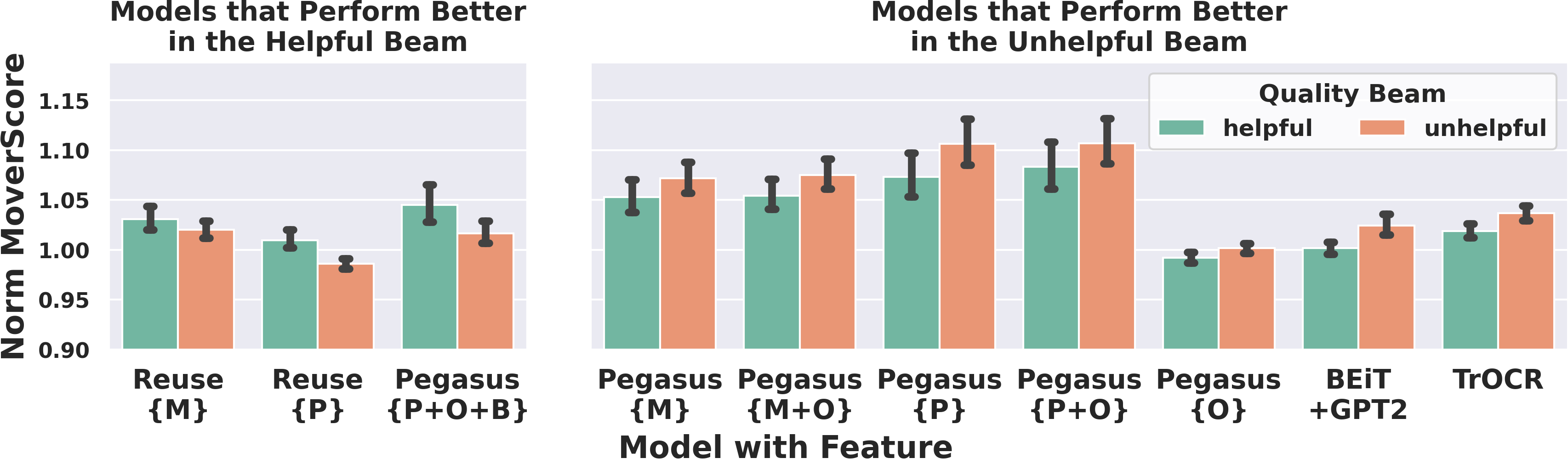}
    \caption{Normalized ROUGE-2 and MoverScore for beams of different quality. Most of the generation models performed better in the unhelpful beam, suggesting that output may be more similar to bad captions. Only the model trained with \textbf{better} captions learned to generate good captions by showing a better score in the helpful beam.}
    \label{fig:quality-beam-rouge-2}
\end{figure*}

\subsection{Challenge 1: Addressing Unreliable Quality of Real-World Data\label{sec:analysis-challenge-1}}

Table~\ref{tab:quality-annotation-stat} shows that \textbf{over 50\% of the author-written captions in arXiv \texttt{cs.CL} papers were deemed unhelpful}.
High unhelpful caption proportion 
may skew evaluation results, particularly for automatic evaluations that compare generated text to human-written captions.
To address this, we evaluated models on different quality beams using the 399 annotated figure captions shown in Table~\ref{tab:quality-annotation-stat}.
The captions were divided into the ``helpful beam'' (184 captions rated \texttt{[Agree]}) and the ``unhelpful beam'' (215 captions rated \texttt{[Disagree]}).

\paragraph{Automatic Evaluation Over Beams of Different Quality.\label{sec:auto-eval-quality-beam}}
To validate the effect of low-quality captions, we re-performed the automatic evaluation for the helpful and unhelpful beam sets.
\Cref{fig:quality-beam-rouge-2} shows the Normalized ROUGE-2 and MoverScore scores for each model in the helpful and unhelpful beam sets.\footnote{In addition, ROUGE-1, ROUGE-L, and BERTScore scores can be found in \Cref{fig:quality-beam-rouge-rest} in \Cref{sec:additional-results}.}
Most models performed better in the unhelpful beam, except Pegasus$_{P+O+B}$, which had better scores in the helpful beam.
Pegasus$_{P+O+B}$ was trained on captions with more than 30 tokens.
This result suggests that improving training data quality, such as by using only longer captions, can positively impact the model's behavior and result in a better generation of helpful captions.

\begin{table}[]
\centering \small

\addtolength{\tabcolsep}{-1mm}
\begin{tabular}{@{}lcccc@{}}
    \toprule
     & \textbf{\#Sample} & \textbf{Peg}$_{P+O}$ & \textbf{Peg}$_{P+O+B}$ & \textbf{Caption} \\ \midrule
    \textbf{Helpful} & 55 & 2.176 & 1.970 & \textbf{1.855} \\
    \textbf{Unhelpful} & 35 & 2.114 & \textbf{1.867} & 2.019 \\ 
    \bottomrule
\end{tabular}
\addtolength{\tabcolsep}{+1mm}
\vspace{-1mm}
\caption{Human ranking results (lower is better) on helpful and unhelpful beams. Pegasus$_{P+O+B}$ received better rankings in the unhelpful beam.}
\label{tab:human-eval-quality-beam}
\vspace{-5mm}
\end{table}

\paragraph{Human Evaluation Over Beams of Different Quality.}
We also re-evaluated human scores for both the helpful and unhelpful beams.
The human evaluation in Section~\ref{sec:human-eval} only covered 90 figures, with 55 in the helpful beam and 35 in the unhelpful beam.
\Cref{tab:human-eval-quality-beam} shows the results.
On average,
\textbf{Pegasus$_{P+O+B}$ (1.867) was ranked better than author-written captions (2.019) in the unhelpful beam, in which}
\textbf{machine-generated captions were preferred by human judges 22 out of 35 times.}
The results suggest that, with careful training data quality control, when author-written captions are not very helpful, machines could potentially generate better captions.

\subsection{Challenge 2: What Constitutes a Good Figure Caption?\label{sec:quality-annotation}}

\begin{table}[t]
    \small
    \center
    \addtolength{\tabcolsep}{-1.5mm}

    \begin{tabular}{@{}lcccc@{}}
    \toprule
                         & \textbf{Image-Text} & \textbf{Visual-Desc} & \textbf{Takeaway} & \textbf{Length} \\ \midrule
    \textbf{Helpfulness} & 0.206        & \textbf{0.523}           & \textbf{0.686}             & \underline{0.383}           \\
    \textbf{Image-Text}  & -            & 0.177           & 0.186             & 0.248           \\
    \textbf{Visual-Desc} & -            & -               & \textbf{0.625}             & \textbf{0.535}\\
    \textbf{Takeaway}    & -            & -               & -                 & \textbf{0.514} \\
    \bottomrule
    \end{tabular}
    \addtolength{\tabcolsep}{+1.5cm}

    \caption{Pearson correlations between different aspects. We used the row scores (five-point Likert scale) to compute the correlation. 
    \textbf{Strong correlation} ($\geq$0.5) and \underline{medium correlation} (0.3 to 0.5) are highlighted.
    Helpfulness is highly correlated with Visual-Description and Takeaway and is moderately correlated with Length.}
    \label{tab-quality-annotation-correlation}
    
\end{table}

We calculated Pearson correlations~\cite{doi:10.1080/00031305.1988.10475524} among the four aspects using raw five-point Likert ratings.
The results are shown in \Cref{tab-quality-annotation-correlation}. 
\textbf{The highest correlation was found between Takeaway and Helpfulness, suggesting that a high-quality caption accurately captures the main message of the figure.}
There were also strong correlations between Helpfulness, Visual-Description, and Takeaway, indicating that a good caption effectively conveys visual information and summarizes the main message.
However,
\Cref{tab:quality-annotation-stat} shows that
only 16.04\% and 18.55\% of the captions described the visual characteristics and the takeaway message, respectively. 

A moderate correlation between Helpfulness and Length supports previous research findings that longer captions are generally more helpful for readers~\cite{hartley2003single,gelman2002let}.

\subsection{Caption Length Distribution}



\begin{figure*}
    \centering
    \begin{subfigure}[b]{0.49\textwidth}
         \centering
         \includegraphics[width=\textwidth]{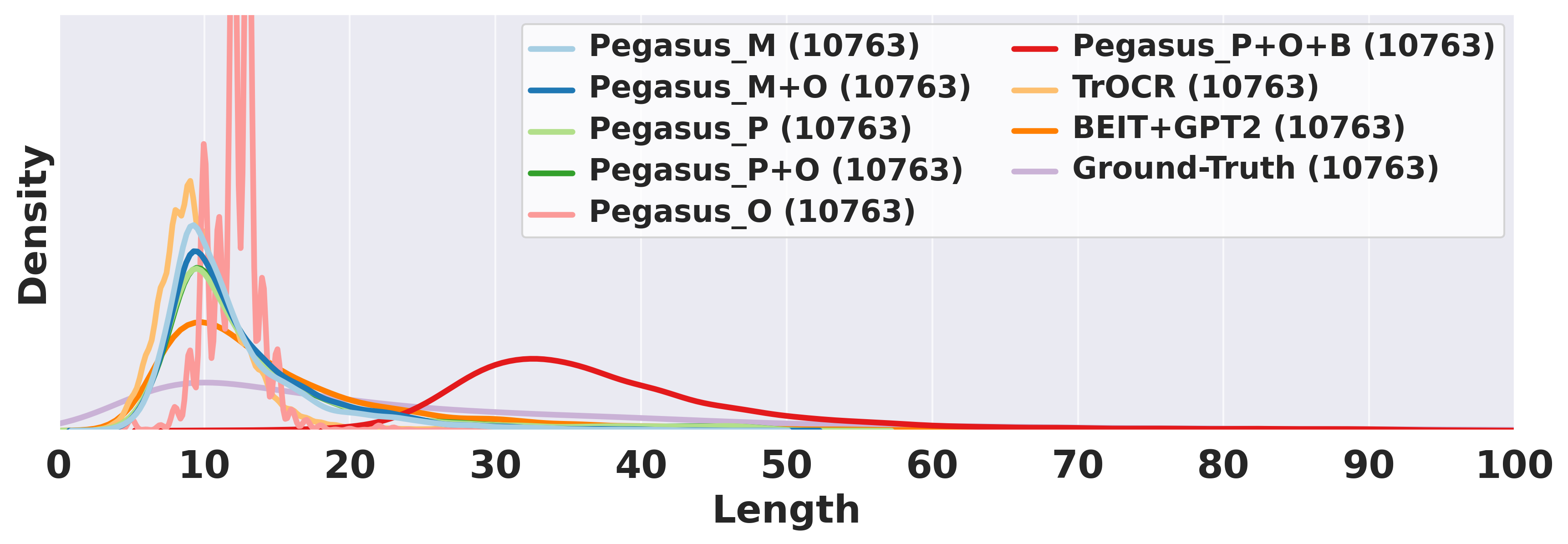}
         \vspace{-6mm}
         \caption{All the examined generative models.}
         \label{fig:length-distribution-systems}
    \end{subfigure}
    \hfill
    \begin{subfigure}[b]{0.49\textwidth}
         \centering
         \includegraphics[width=\textwidth]{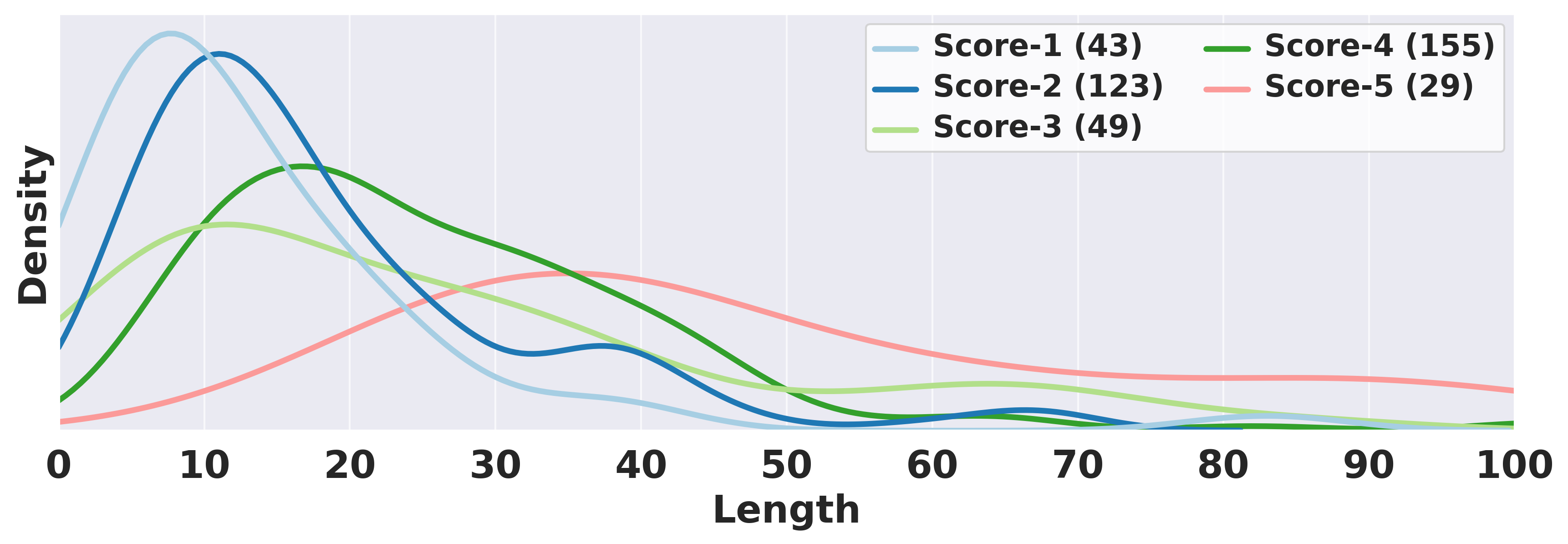}
         \vspace{-6mm}
         \caption{Helpfulness scores in 5-point Likert scale.}
         \label{fig:length-distribution-score-list}
    \end{subfigure} \\
    \begin{subfigure}[b]{0.49\textwidth}
         \centering
         \includegraphics[width=\textwidth]{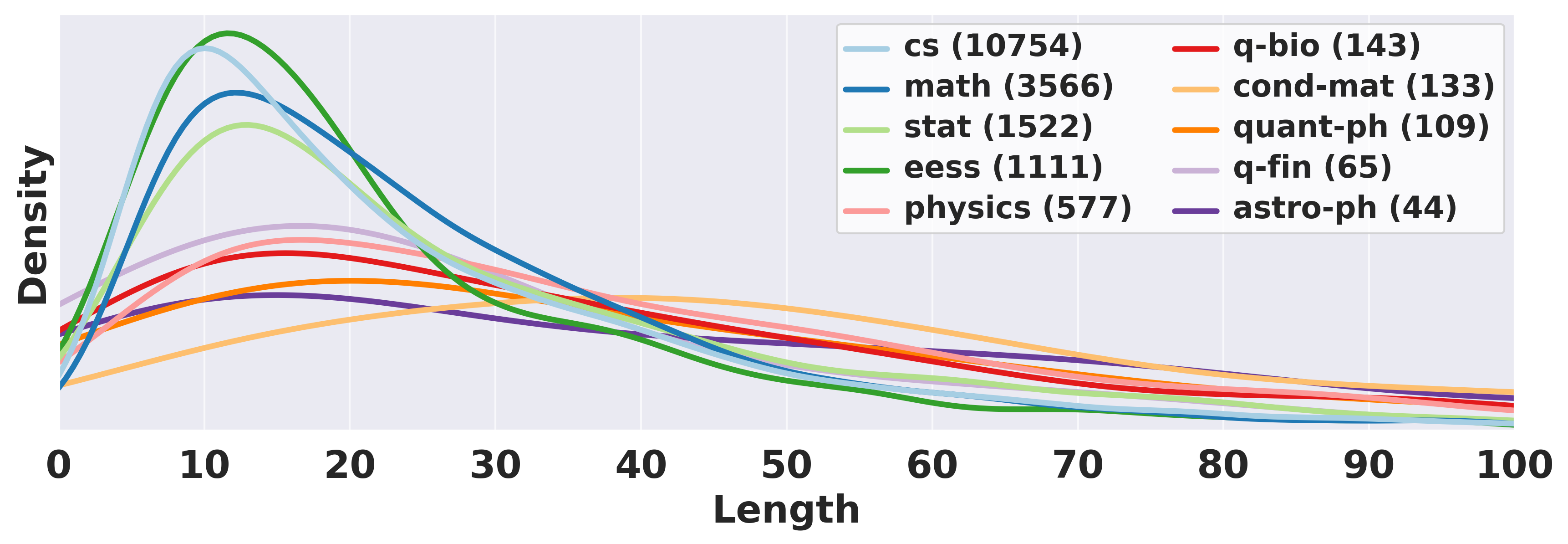}
         \vspace{-6mm}
         \caption{Top 10 frequent arXiv categories.}
         \label{fig:length-distribution-global-category}
    \end{subfigure}
    \hfill
    \begin{subfigure}[b]{0.49\textwidth}
         \centering
         \includegraphics[width=\textwidth]{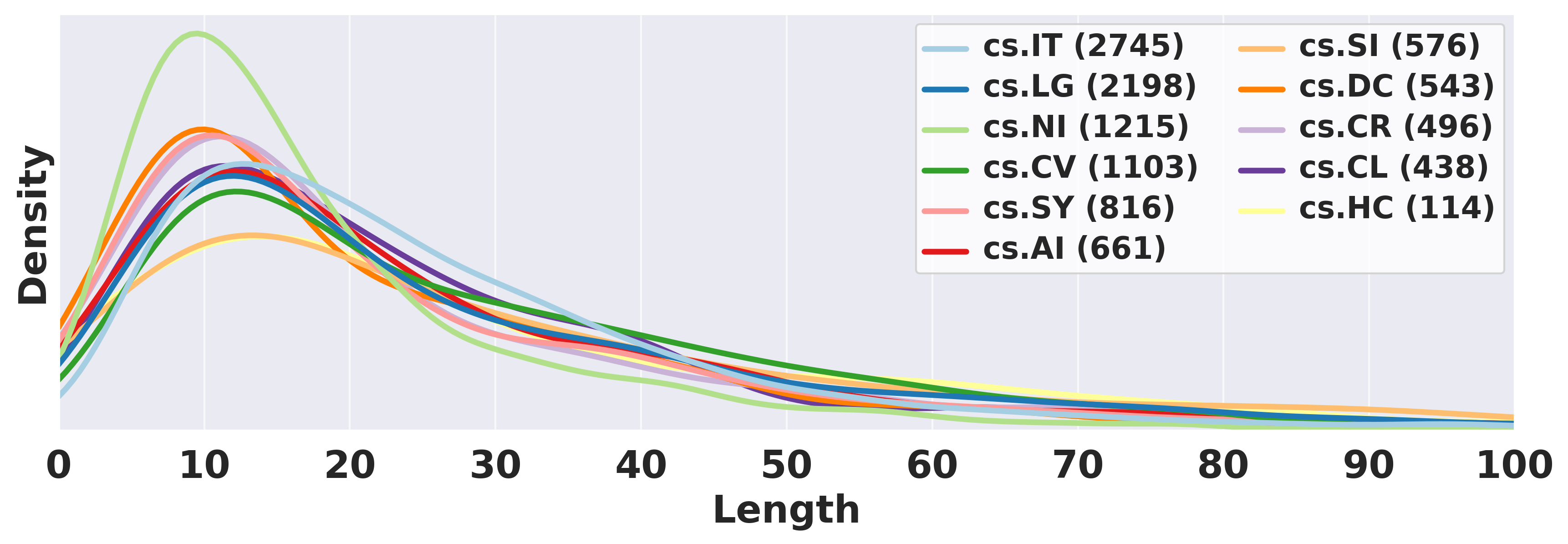}
         \vspace{-6mm}
         \caption{Top 10 frequent arXiv CS subcategories and cs.HC.}
         \label{fig:length-distribution-cs-category}
    \end{subfigure}
    \caption{The KDE plot illustrates diverse caption length distributions among models, the relationship with helpfulness, and variations across arXiv categories. The sample numbers are included in the legend.}
    \label{fig:length-distribution}
    \vspace{-4mm}
\end{figure*}

Throughout this work's development, the length of captions emerged as a consistent issue.
Despite existing literature indicating the benefits of longer captions for readers~\cite{hartley2003single,gelman2002let}, space limitations often leave authors with no option but to craft shorter captions.
To shed some light on this aspect and offer insight for future research, we analyzed the lengths of both author-created and machine-generated captions.
We used Kernel Density Estimate (KDE) plots to investigate the distribution of caption lengths across different models and domains.
As shown in \Cref{fig:length-distribution-systems}, the majority of models demonstrate a common peak at 10 tokens, while
Pegasus$_{P+O+B}$ presents a significant deviation with a peak near 30 tokens.
%
\Cref{fig:length-distribution-score-list} presents the distribution of helpfulness scores, 
derived from quality annotation data (see \Cref{sec:quality-annotation}).
Captions rated with a maximum helpfulness score of 5 show a peak at 35 tokens.
We can also see a clear shift in caption length with higher scores.
In \Cref{fig:length-distribution-global-category}, we dug into the top 10 category taxonomy from arXiv.
This figure suggests that a higher portion of the captions in \texttt{cs}, \texttt{math}, \texttt{stat}, and \texttt{eess} are shorter (10 tokens); while the rest of the categories (\texttt{cond-mat}, \texttt{quant-ph}, \texttt{q-bio}, \emph{etc}) 
have higher probabilities for longer captions.
%
%
However, within the \texttt{cs} domain (\Cref{fig:length-distribution-cs-category}), the top 10 subcategories do not show significant differences regarding caption length distribution.

\section{Discussion\label{sec:discussion}}


\begin{table}[t]
    \centering
    \small
    
    \begin{tabular}{@{}lccc@{}}
        \toprule
        \textbf{Information} & \textbf{Image-Text} & \textbf{Visual-Desc} & \textbf{Takeaway}  \\ \midrule
        \textbf{\#Tokens} & 0.181 & 0.428 & 0.357 \\
        \textbf{Percentage} & 0.099 & 0.279 & 0.210 \\ \bottomrule
    \end{tabular}

    \caption{Correlations between the amount of missing information from Paragraph and the quality aspects. The missing information is related to visual descriptions and takeaway messages.}
     \vspace{-2mm}    
    \label{tab:missing-alignment}
    
\end{table}

\paragraph{Is Text Really All You Need?}
Our results demonstrate that summarizing figure-mentioning paragraphs is sufficient to generate captions, as shown by the similar scores of Pegasus$_{P}$ and Pegasus$_{P+O}$ in Table~\ref{tab:result-summarization}.
Adding OCR had limited impact.
Furthermore,
in a recent study of scientific figure captioning conducted by Yang {\em et al.}~\cite{scicap-plus}, the best-performing model only considered figure-mentioning paragraphs and OCR tokens-- note that their OCR tokens were visual features-- without taking the figure's imagery into account. 
These results raise an interesting question: 
Do we need visual information at all? What for?
The token alignment study (\Cref{sec:motivating-analysis}) showed that 75.19\% of the caption information could be found in the Paragraphs, meaning 24.81\% of the information was missing.
Understanding this missing information could help improve the models' performance.
Thus, we calculated the correlation between the amount of missing information and three aspect ratings (image-text, visual-description, and takeaway) in the quality annotation data (\Cref{sec:quality-annotation}).
The missing information was quantified as the number or percentage of tokens without aligning to any tokens in figure-mentioning paragraphs.
\Cref{tab:missing-alignment} demonstrates a positive correlation between the extent of missing information and visual descriptions and takeaway messages.
This suggests that incorporating visual descriptions (\eg, ``dashed line,'' ``red line'') is key to enhancing performance by filling in the gaps in information not covered by the article's text.
Furthermore, the strong correlation between Helpfulness and Visual-Description in~\Cref{tab:quality-annotation-stat} also indicates that including image information is necessary for writing good captions.
It should be noted that OCR is only capable of capturing image texts (\eg, labels, legends) and not visual element information (\eg, ``dashed line'').
A promising future direction is developing a multimodal model that can effectively incorporate both image and text.


\paragraph{What is the \textit{Best} Length for Captions?}
Our research indicates that filtering shorter captions can facilitate the generation of more helpful captions.
However, the resulting captions tend to be longer than usual, as shown in the Pegasus$_{P+O+B}$ shift to the right in \Cref{fig:length-distribution-systems}. 
This raises a question: Is it fair to compare short and long captions on usefulness, given that longer captions inherently contain more information?
While our automatic evaluation addressed this by implementing length normalization, our human evaluations and quality annotations did not specifically instruct the annotators to consider caption lengths.
Nevertheless, we argue that even if we asked annotators to consider caption lengths while identifying helpful captions, 
the ``ideal'' caption length would differ among annotators due to multiple factors.
For example, as shown in \Cref{fig:length-distribution-global-category}, the length distributions of captions vary across domains.
%
The low inter-agreement from our human evaluation (see \Cref{sec:human-eval}) also suggests that personal preferences could influence ideal caption length~\cite{lundgard2021accessible}.
%
Moreover, the ideal length could also be dictated by the context: writers might favor shorter captions due to page constraints, while readers might prefer longer but informative ones~\cite{stokes2022text,sun-etal-2019-compare}.
To tackle this issue, a potential future direction could be enabling models to generate captions of diverse lengths to suit different users and contexts.

\section{Conclusion and Future Work}
This work presented a new perspective on automatic figure captioning, demonstrating that a language-based approach, \ie, summarizing figure-referring paragraphs, can outperform conventional vision-based methods.
Our analysis further showed many unhelpful captions in arXiv papers, highlighting data quality's impact on captioning performance.
This work lays the groundwork for further research, including exploring new data selection, revision, and augmentation strategies to mitigate the effects of low-quality data, developing new evaluation methods, and creating more robust models that better handle noisy data.
We also aim to expand the technology's scope to cover a wider variety of figures and article types.

\section*{Acknowledgements}
Thanks to the anonymous reviewers for their constructive feedback.
This work is supported by Adobe's gift funds and the seed funds from Pennsylvania State University's College of Information Sciences and Technology (IST).

\section*{Limitations}
Although our proposed methods have been shown to be effective, we are aware of several limitations.
First, our approach requires mentions in order to produce captions, but it is not always easy to automatically identify the mentions for a given figure in real-world data. 
There were 18.81\% of figures in the original \oldDataset that did not have any identified mentions, which we excluded from this work.
Many factors contributed to the gap, including errors caused by upstream components such as image extraction or image type classification (\eg, table), unexpected figure index formats (\eg, ``Figure VIII'', ``Figure C·1'',``Fig.Fig. 4(b)''), PDF parsing errors, or the figure never being mentioned in the paper.
Second, our method uses texts instead of images as the primary information source, so, naturally, it inherits all the constraints of text. 
Our method can not capture any visual element in the figure that the text never mentioned; it struggles when the text is poorly written.
Finally, this paper focused on non-compound line charts in arXiv papers; the human evaluation only focused on NLP papers. 
More research is needed to examine the generalizability.

\section*{Ethics Statement}
We consider the proposed technology to impose little risk to readers, as it only summarizes what has already been presented in the paper.
However, when the generated caption contains some inaccurate information, it could mislead readers.
Furthermore, the proposed technology has the nature of neglecting visual content, which might have an impact on the accessibility of figure captions.

\bibliography{bib/anthology,bib/custom,bib/figure_caption,bib/text_summarization}
\bibliographystyle{acl_natbib}

\clearpage

\appendix

\section{Token Overlap Study}
\label{sec:token-overlap-study} 
This is the additional study we conducted to support \Cref{sec:motivating-analysis}.
We computed n-gram precision scores (BLEU-4) for mentions and captions
extracted from different settings.
For mentions, we included 
{\em (i)} First Mention, the sentence that first mentions the figure in the paper;
{\em (ii)} Random Mention, a randomly selected sentence among all mentions;
and 
{\em (iii)} Random Sentence, a randomly selected sentence from the paper.
We also included one or two following sentences, as the surrounding context may contain relevant information for the figure. 
For captions, we examined: 
{\em (i)} First Caption, the first sentence of the caption; 
and {\em (ii)} Whole Caption, all the sentences in the caption. 
The results are shown in \Cref{tab:analysis_caption-mention}. 
The extremely low BLEU-4 score for the Random Sentence baseline (First: 0.01, Whole: 0.01) indicates 
that a randomly selected sentence has very limited information related to the caption.
In contrast, the results for First Mention (First: 9.39, Whole: 10.54) and Random Mention (First: 9.15, Whole: 10.28) show 
the presence of significantly more information relevant to the caption.
All scores in First Mention are slightly better than the corresponding ones in Random Mention, 
suggesting that writers tend to give more detail when they first introduce the figure.




\section{Data Preprocessing Details}
\label{sec:appendix-data-preprocessing}
We describe the detailed data preprocessing steps here as supplementary materials for \Cref{sec:data}.

\paragraph{Dataset Resplit.}
\oldDataset was originally created for vision-to-lan\-guage tasks, and
we needed new train/val/test splits for our work. 
As figures do not overlap, different figures in the same paper can be assigned to different data splits in \oldDataset.
However, papers' texts overlap more easily and can be problematic for text-summarizing tasks. 
We resplit \oldDataset to make sure no paper had figures from different data splits, 
and we excluded figures without any identified Mentions. 
As a result, the [train/val/test] sets used in this work had [86,825/10,833/10,763] figures 
sourced from [48,603/6,055/6,053] papers, respectively.\footnote{The Paragraph+OCR-Better model, which used captions with more than 30 tokens, was trained on only 27,224 samples.}

\paragraph{OCR.}
The OCR texts were extracted from all figures using EasyOCR~\cite{easyocr}.
The output from EasyOCR included the OCR texts along with their bounding boxes.
In order to incorporate the OCR texts into our models, we concatenated them with the sequence of coordinates obtained by traversing the bounding boxes from left to right and then from top to bottom.

\paragraph{Representative.}
It is worth noting that we manually verified 399 figures (the set used in \Cref{sec:analysis}) and found that 81.2\% (324/399) was published at academic conferences, and 51.9\% (207/399) were at ACL Anthology, IEEE, or ACM, suggesting that the data is representative.

\begin{table}[t]
    \centering \small
    
    \begin{tabular}{lrr}
    \toprule
    \multicolumn{1}{c}{\multirow{2}{*}{\textbf{Setting}}} & \multicolumn{2}{c}{\textbf{Maximum Length}} \\ \cmidrule{2-3}
    & \textbf{Source} & \textbf{Target} \\ \midrule
    \textbf{Mention}        & 128 & 100 \\
    \textbf{Mention+OCR}    & 256 & 100 \\ 
    \textbf{Paragraph}      & 512 & 100 \\
    \textbf{Paragraph+OCR}  & 640 & 100 \\
    \textbf{Paragraph+OCR-Better} & 640 & 140 \\
    \textbf{OCR}            & 128 & 100 \\
    \bottomrule
    \end{tabular}
    \vspace{-2mm}
    \caption{Maximum length configuration for the text summarization models.}
    \label{tab:length-setting}
    \vspace{-2mm}
\end{table}

\begin{table}[t]
    \centering \small
    \addtolength{\tabcolsep}{-1.4mm}
    
    \begin{tabular}{@{}lrrrrrrrrr@{}}
    \toprule
     \multicolumn{1}{@{}c}{\textbf{Caption}} & \multicolumn{3}{c}{\textbf{FST Mention}} & \multicolumn{3}{c}{\textbf{RDM Mention}} & \multicolumn{3}{c@{}}{\textbf{RDM Sentence}} \\ \midrule
     
    \textbf{Context} & \multicolumn{1}{c}{\textbf{+0}} & \multicolumn{1}{c}{\textbf{+1}} &  \multicolumn{1}{c}{\textbf{+2}} & \multicolumn{1}{c}{\textbf{+0}} & \multicolumn{1}{c}{\textbf{+1}} & \multicolumn{1}{c}{\textbf{+2}} & \multicolumn{1}{c}{\textbf{+0}} & \multicolumn{1}{c}{\textbf{+1}} & \multicolumn{1}{c}{\textbf{+2}} \\ \midrule
    
    \textbf{First} & 9.39 & 6.25 & 4.91 & 9.15 & 6.09 & 4.78 & 0.01 & 0.59 & 0.54 \\
    \textbf{Whole} & \textbf{10.54} & 8.08 & 6.96 & 10.28 & 7.92 & 6.83 & 0.01 & 0.80 & 0.76 \\ 
    \bottomrule
    
    \end{tabular}
    \addtolength{\tabcolsep}{+1.4mm}

    \caption{N-gram matching (BLEU-4) between captions and mentions of each figure. First and Whole refer to the first sentence of the caption and the whole caption, respectively. Context means the number of the following sentences included. First Mention was better than Random Mention in the corresponding settings, suggesting that writers may give more details when first introducing the figure.}
    \label{tab:analysis_caption-mention}
    
\end{table}

\section{Training and Decoding Details}
\label{sec:appendix-model-training-details}
We describe the model training details and the decoding configuration used in \Cref{sec:training-setup}.

\paragraph{Training Details for Text Summarization Models.}
We fine-tuned Pegasus\footnote{We used \texttt{google/pegasus-arxiv}.} for the text-summarization task using HuggingFace's implementation~\cite{wolf-etal-2020-transformers}. 
All the models shared the same training hyper-parameters except maximum text length, 
as the data varies in all the examined settings. 
The maximum source length and target length were set to 
(\textit{i}) fully cover at least 95\% of text without truncation and 
(\textit{ii}) be able to fit into the machine. 
We show the length configuration in \Cref{tab:length-setting}. 
Other hyper-parameters used for training were 
batch size = 32, learning rate = 5e-5 with a linear decay scheduler, and number of training epochs = 200. 
We evaluated the model every five epochs, and the one with the highest ROUGE-2 score was kept for testing~\cite{liu-liu-2021-simcls,zhong-etal-2020-extractive,xu-etal-2020-discourse}. 
All models were trained with an NVIDIA A100 GPU. Each model took one to three days to train.


\paragraph{Training Details for Vision-to-Language Models.}
Two vision-to-language models were fine-tuned using HuggingFace:
(\textit{i}) a sequence-to-sequence model using BEiT\footnote{We used \texttt{microsoft/beit-large-patch16-384}.} and GPT-2\footnote{We used \texttt{gpt2-large}.}
and (\textit{ii}) TrOCR.\footnote{We used \texttt{microsoft/trocr-large-printed}.}  
The hyperparameters used for training were maximum target length = 100, 
learning rate = 2e-5 with a linear warmup (one epoch), and linear decay scheduler. 
Batch sizes were 32 and 64, respectively.
The models were trained using AdamW~\cite{loshchilov2018decoupled}, with weight decay = 1e-4 for 100 epochs. 
We evaluated the model every epoch and kept the one with the highest ROUGE-2 score~\cite{liu-liu-2021-simcls,zhong-etal-2020-extractive,xu-etal-2020-discourse}. 
The model was trained with an NVIDIA A100 GPU for two days.

\paragraph{Decoding.}
For all generation models, captions were decoded using the beam sampling strategy, 
with beam size = 5, temperature = 0.8, top-k = 100, repetition penalty = 3.0, 
minimal length = 10, and maximum length = 100.

\section{Interfaces\label{sec:app-interface}}

\begin{figure*}[t]
    \centering
    \includegraphics[height=.95\textheight, frame]{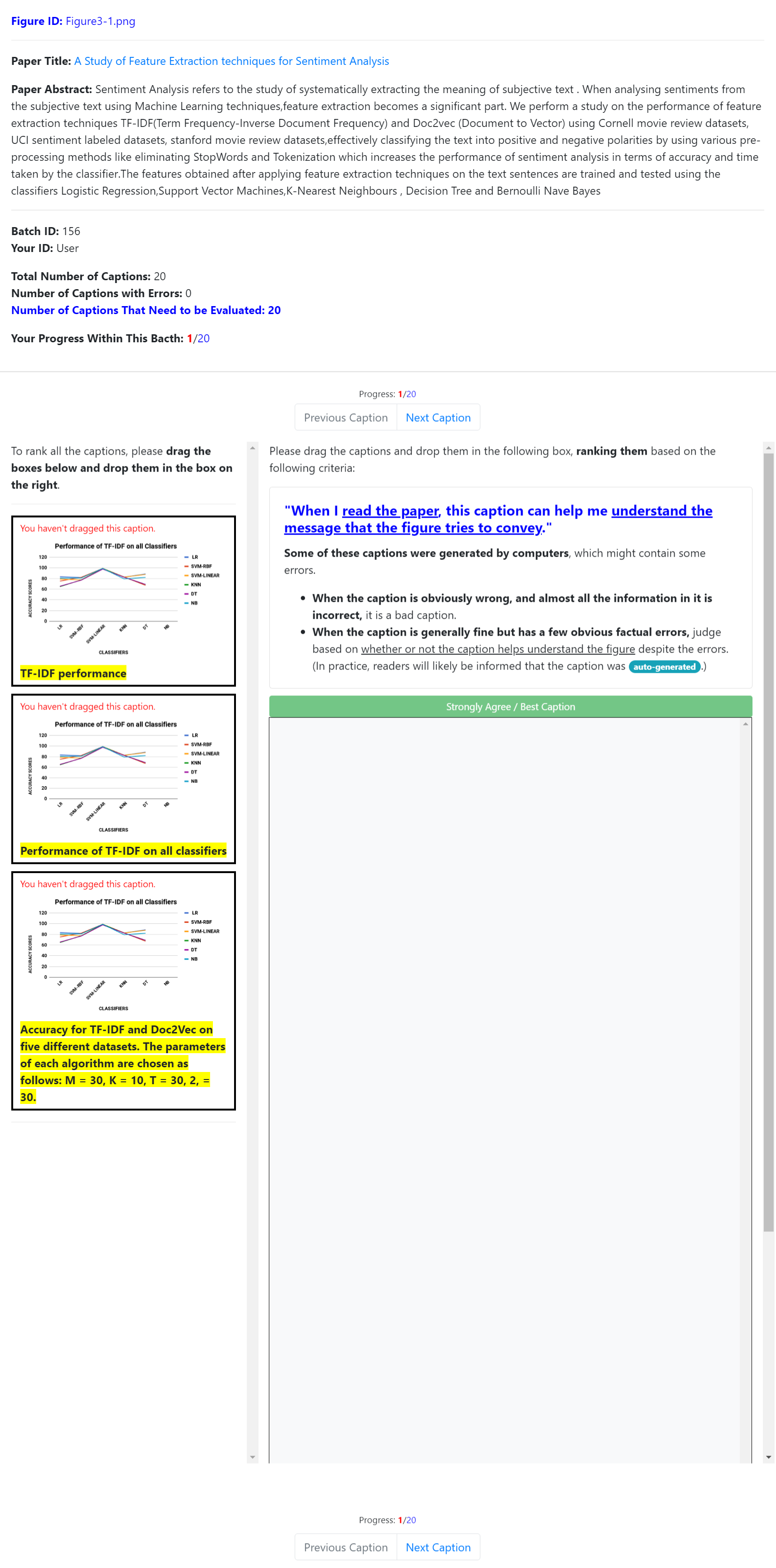}
    \caption{The interface the human judges used to rank the captions (see Section~\ref{sec:human-eval}).}
    \label{fig:ui-compare}
\end{figure*}

\Cref{fig:ui-compare} shows the interface the human judges used to rank the captions (see Section~\ref{sec:human-eval}).
The paper's title (without linking to the paper's URL) and abstract are shown. 
The human judges can drag the captions (each displayed with the figure) on the left pane and drop them to the right pane to rank them.
The initial display order of the captions is randomized on the interface.
We did not display the paper's PDF or link to the paper's URL to prevent human judges from biasing toward the author-written captions.

\begin{figure*}[t]
    \centering
    \includegraphics[height=.95\textheight, frame]{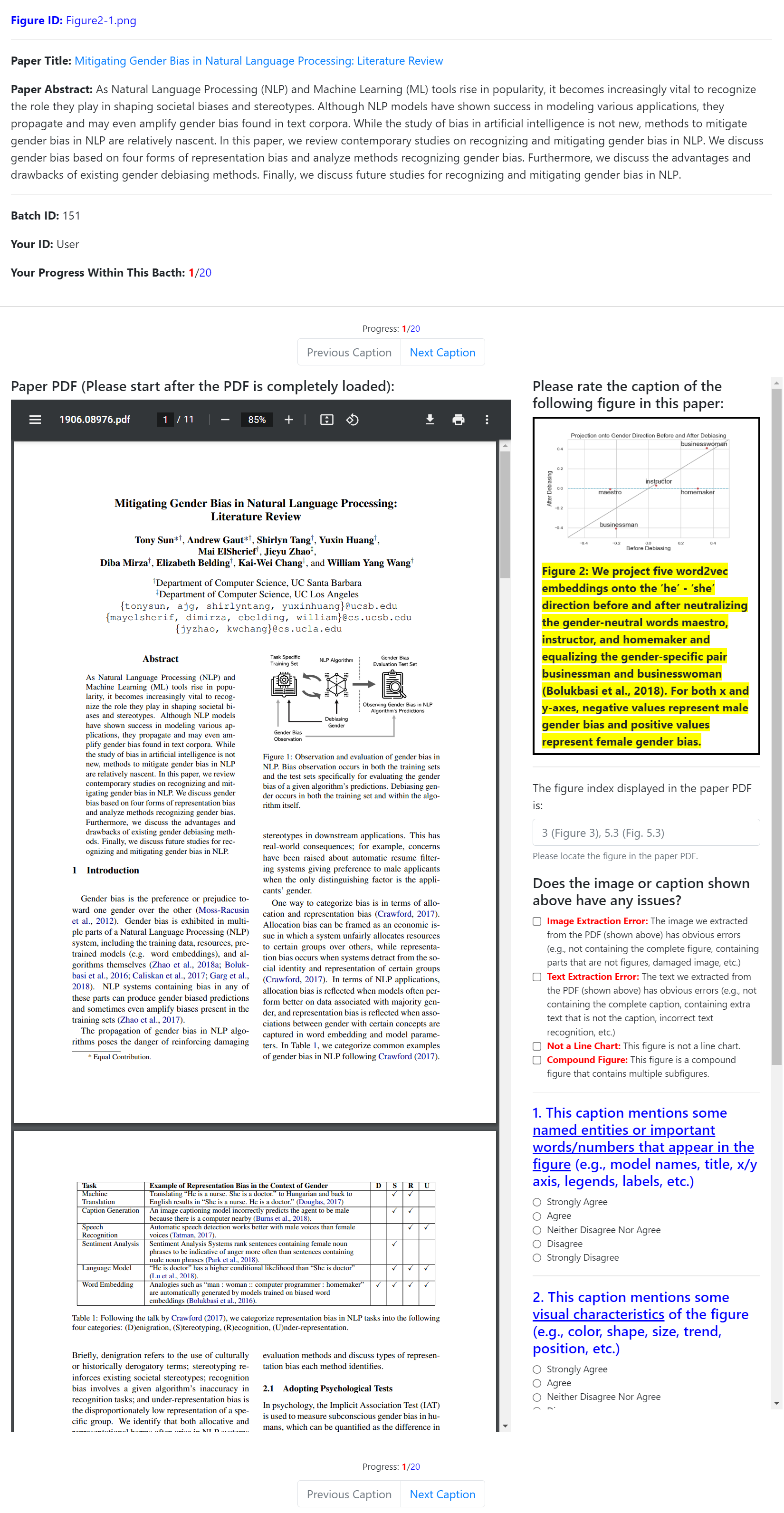}
    \caption{The interface we used to rate the usefulness of captions (see Section~\ref{sec:quality-annotation}).}
    \label{fig:ui-rating}
\end{figure*}

\Cref{fig:ui-rating} shows the interface we used to rate the usefulness of captions (see Section~\ref{sec:quality-annotation}.)
The title (with a hyperlink to the paper's URL), abstract, and the PDF file of the paper were shown, alongside the target figure's image/caption and all the questions.
We displayed the paper's PDF to help raters make more informed decisions on the caption quality.

\section{Additional Experimental Results\label{sec:additional-results}}
In this section, we show all the additional experimental results mentioned in the experiment and analysis.

\paragraph{Normalization Scores.}
\Cref{fig:length-performance-rouge1,fig:length-performance-rougel,fig:length-performance-wms,fig:length-performance-bertscore} shows the relationship between generation text length and performance (ROUGE-1, ROUGE-L, MoverScore, and BERTScore). The random lines indicate that the text length and the performance are not independent, suggesting that normalization over text length is needed.
\Cref{tab:random-scores} shows the corresponding random scores for each of the metrics used in \Cref{tab:result-summarization}.

\begin{figure*}[t]
    \centering
    \includegraphics[width=0.99\linewidth]{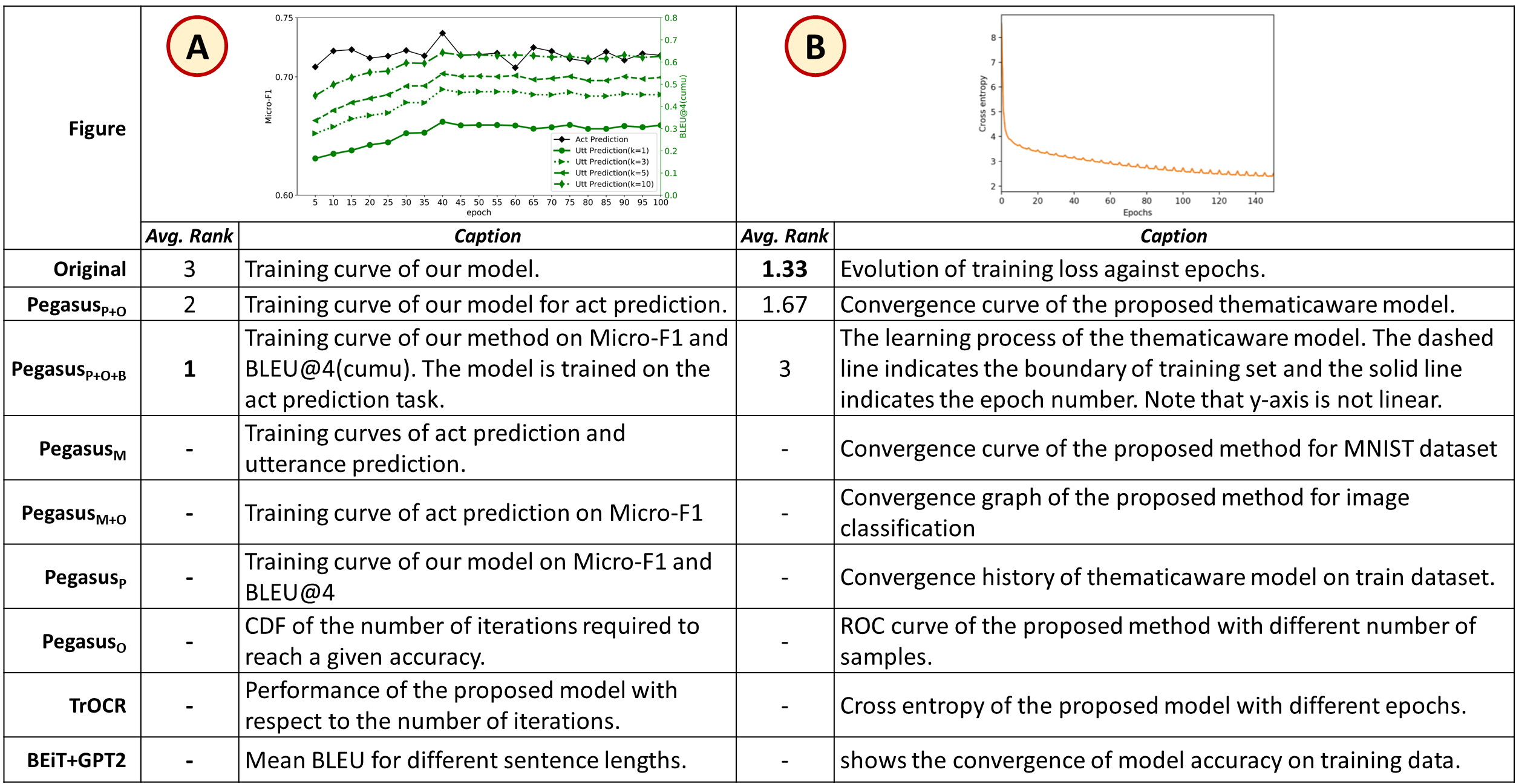}
    \vspace{-2mm}
    \caption{Example output from text-summarization models. The top three rows were included in the human evaluation. Three human judges ranked the three captions for each figure (from 1 to 3 where lower is better). (A) Pegasus$_{P+O+B}$ was preferred, as it provided more details about the figure. [Image/caption source:~\cite{jiang2019dialogact2vec}] (B) Pegasus$_{P+O+B}$ generated a bad caption by introducing many obvious factual errors. [Image/caption source:~\cite{wang2019theme}]}
    \vspace{-2mm}
    \label{fig:good-bad-example}
\end{figure*}

\begin{figure*}[t]
    \centering
    \includegraphics[width=0.48\linewidth]{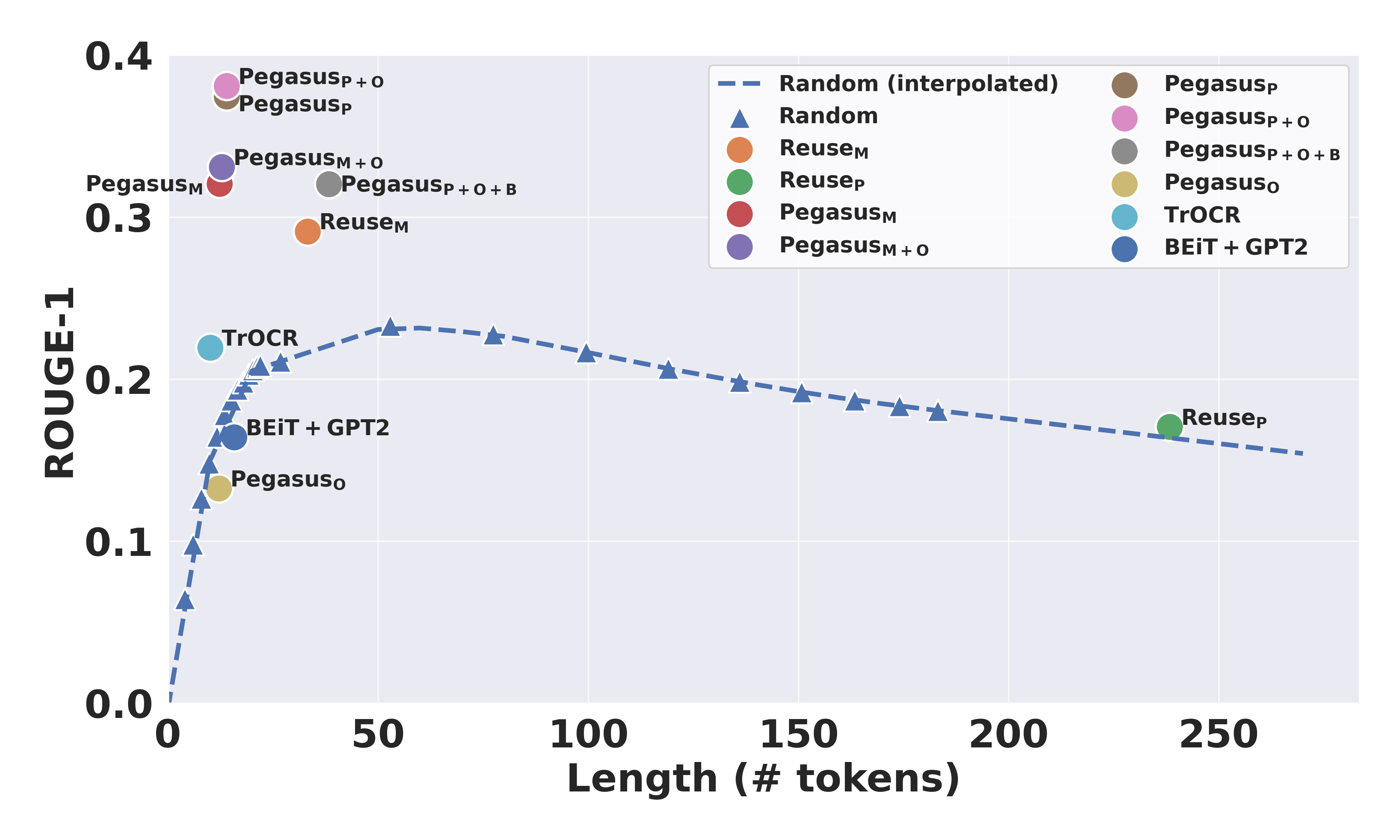}
    \includegraphics[width=0.48\linewidth]{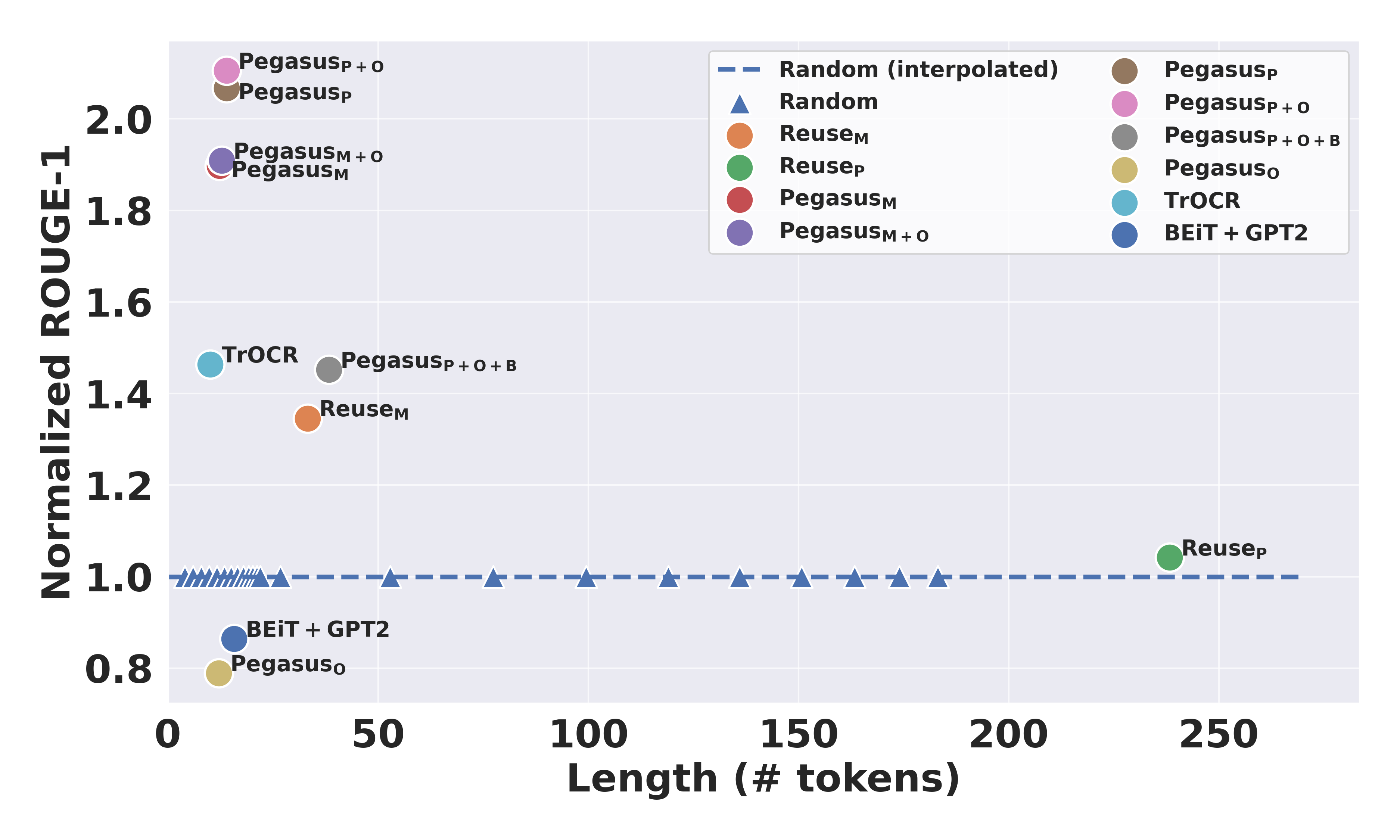}
    \vspace{-3mm}
    \caption{The relationship between average text length and ROUGE-1. When the generated text is shorter than 50 tokens, longer texts generally result in a higher ROUGE-1 score.} 
    \label{fig:length-performance-rouge1}
\end{figure*}

\begin{figure*}[t]
    \centering
    \includegraphics[width=0.48\linewidth]{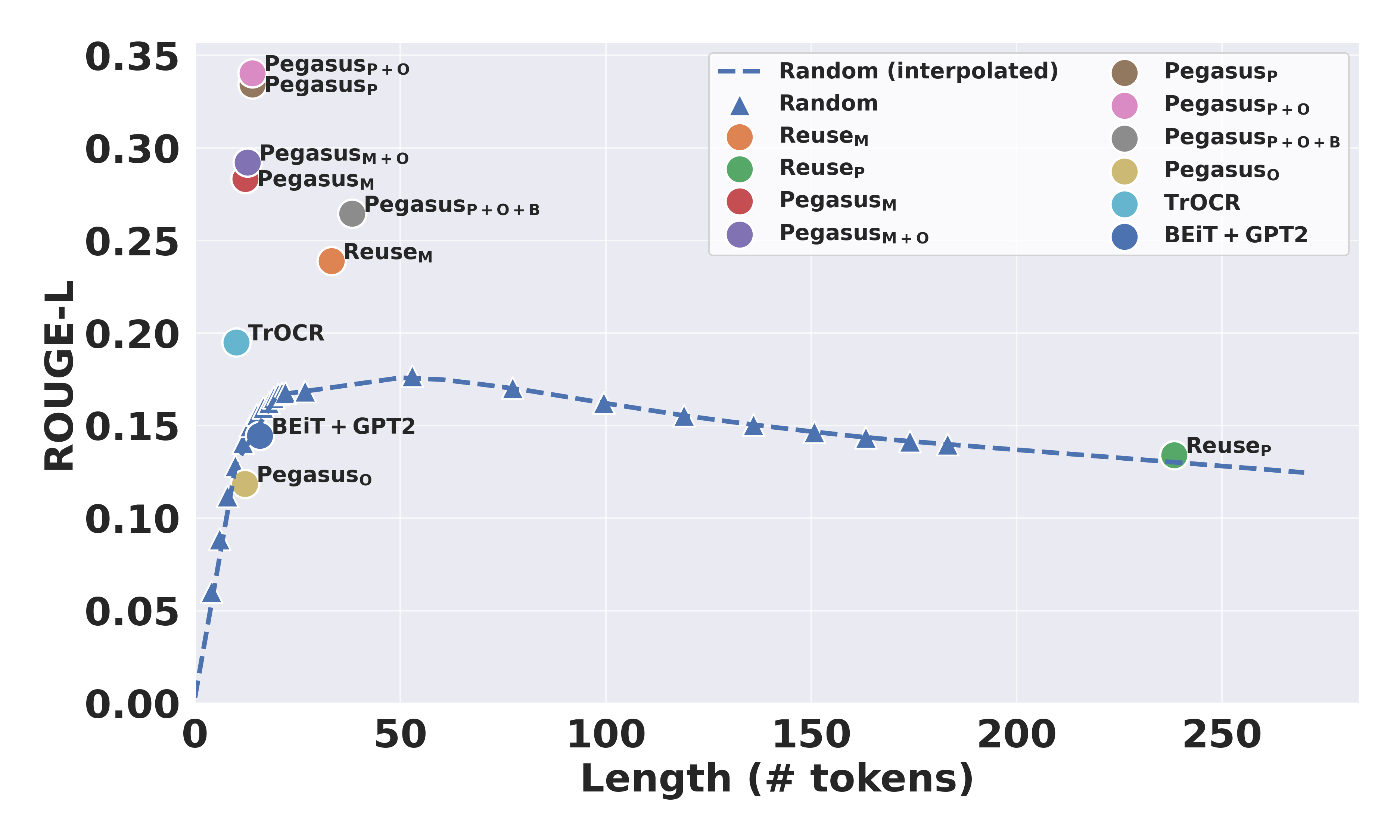}
    \includegraphics[width=0.48\linewidth]{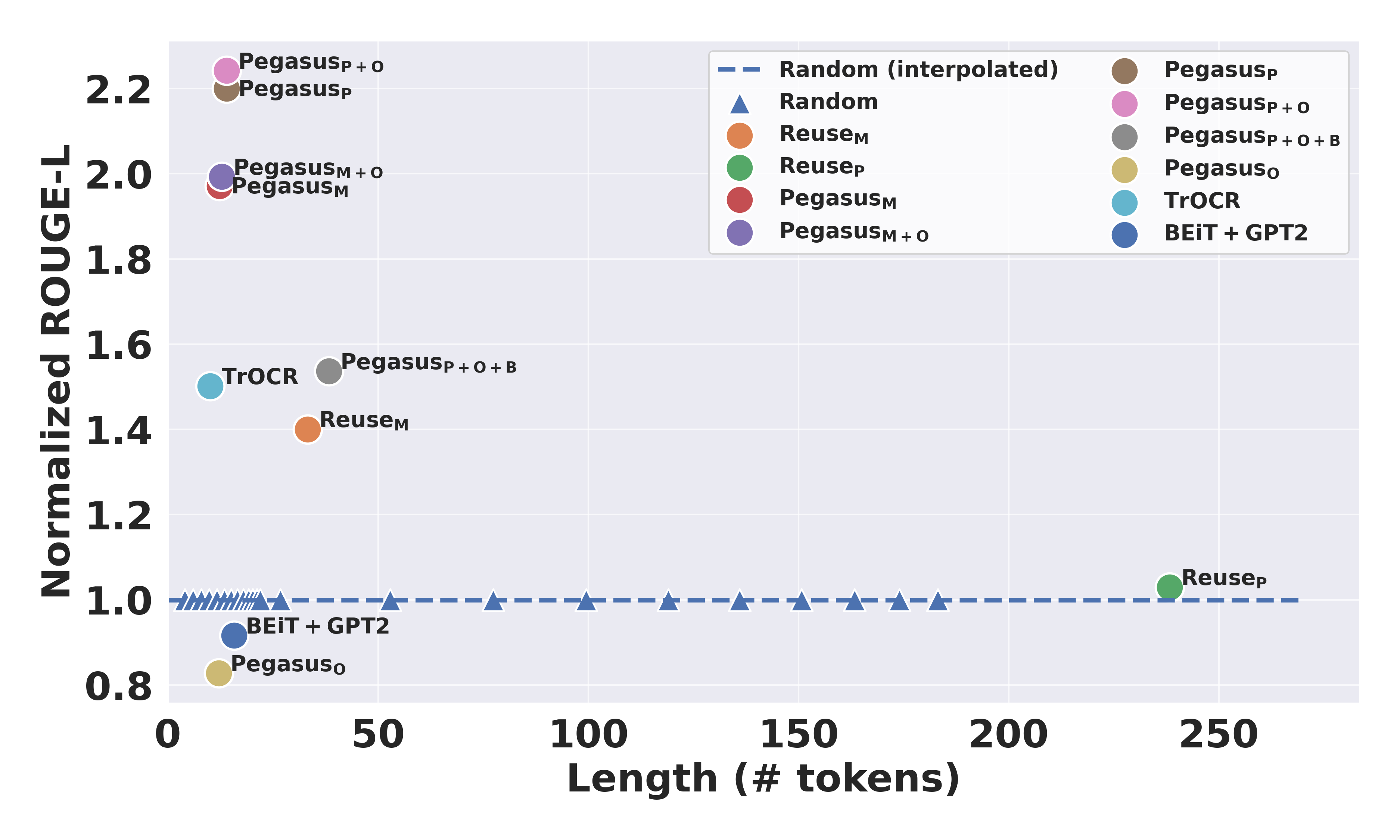}
    \vspace{-3mm}
    \caption{The relationship between average text length and ROUGE-L. When the generated text is shorter than 50 tokens, longer texts generally result in a higher ROUGE-L score.} 
    \label{fig:length-performance-rougel}
\end{figure*}

\begin{figure*}[t]
    \centering
    \includegraphics[width=0.48\linewidth]{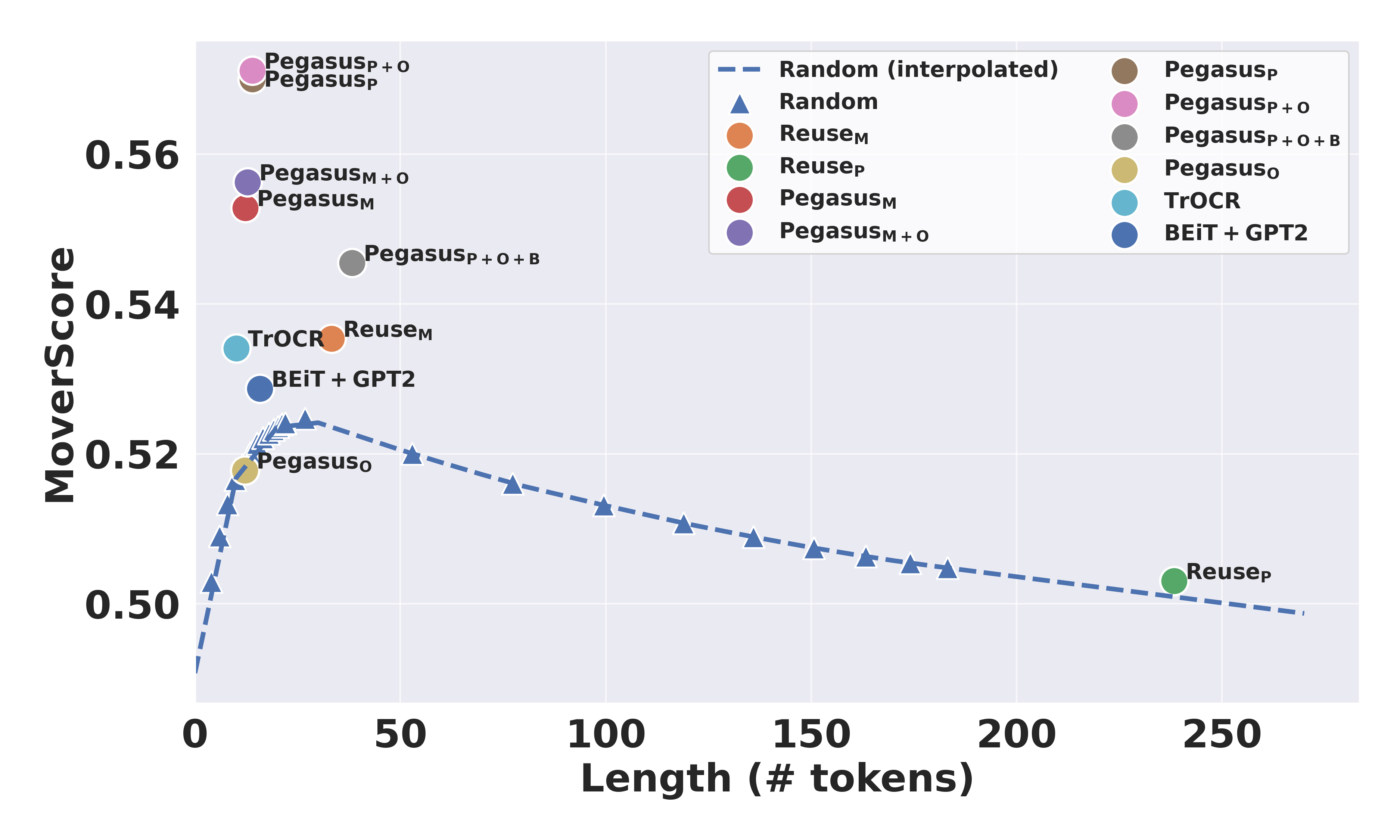}
    \includegraphics[width=0.48\linewidth]{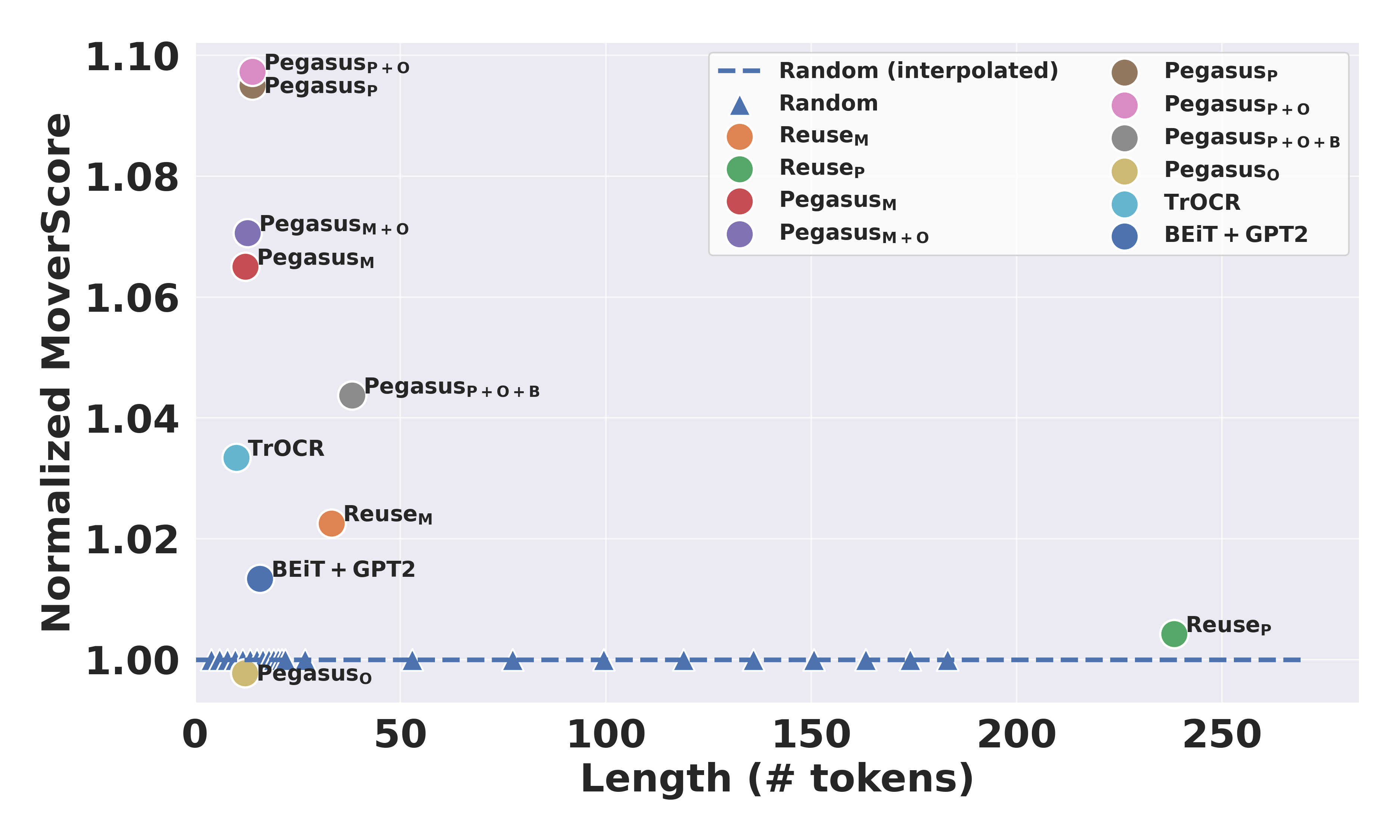}
    \vspace{-3mm}
    \caption{The relationship between average text length and MoverScore. When the generated text is shorter than 30 tokens, longer texts generally result in a higher MoverScore score.} 
    \label{fig:length-performance-wms}
\end{figure*}

\begin{figure*}[t]
    \centering
    \includegraphics[width=0.48\linewidth]{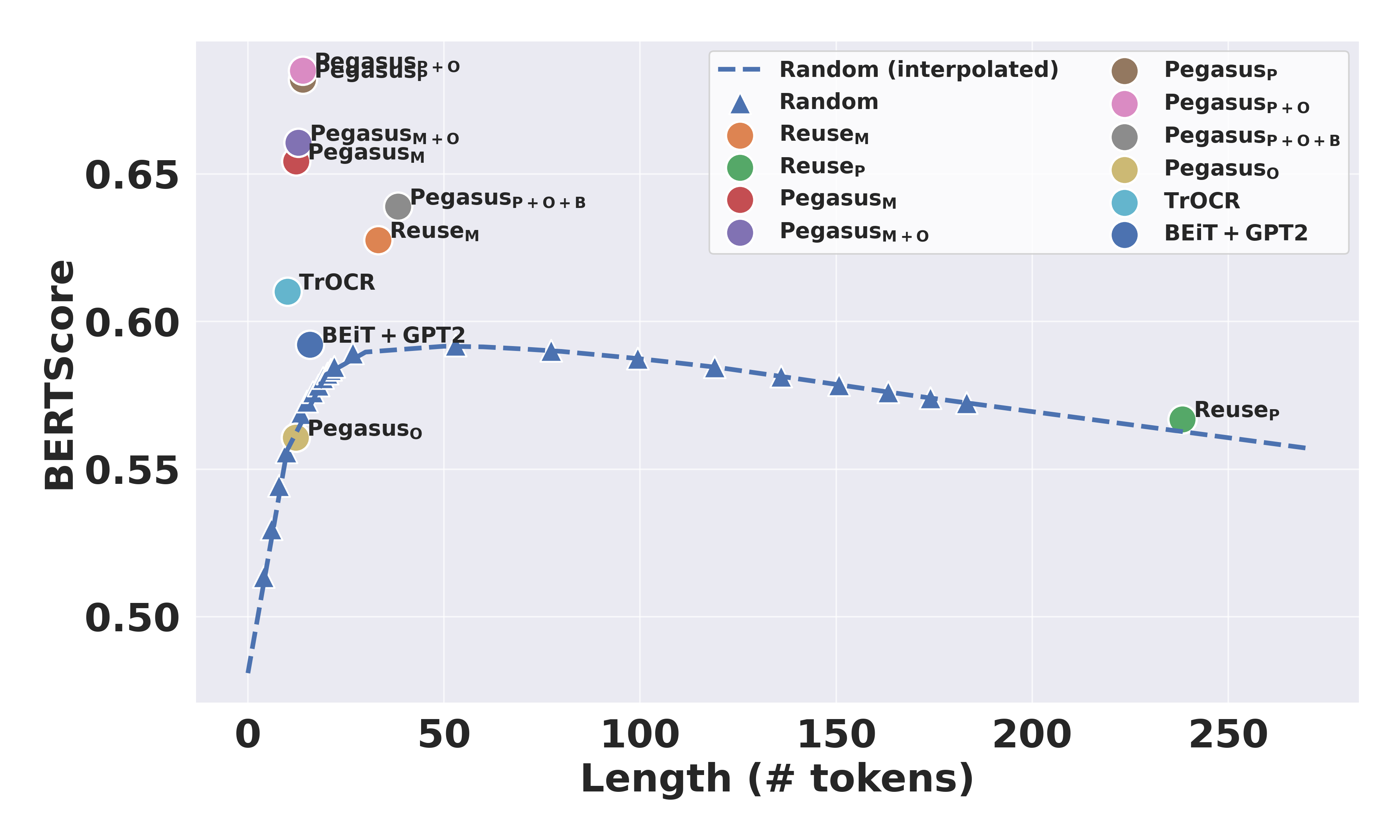}
    \includegraphics[width=0.48\linewidth]{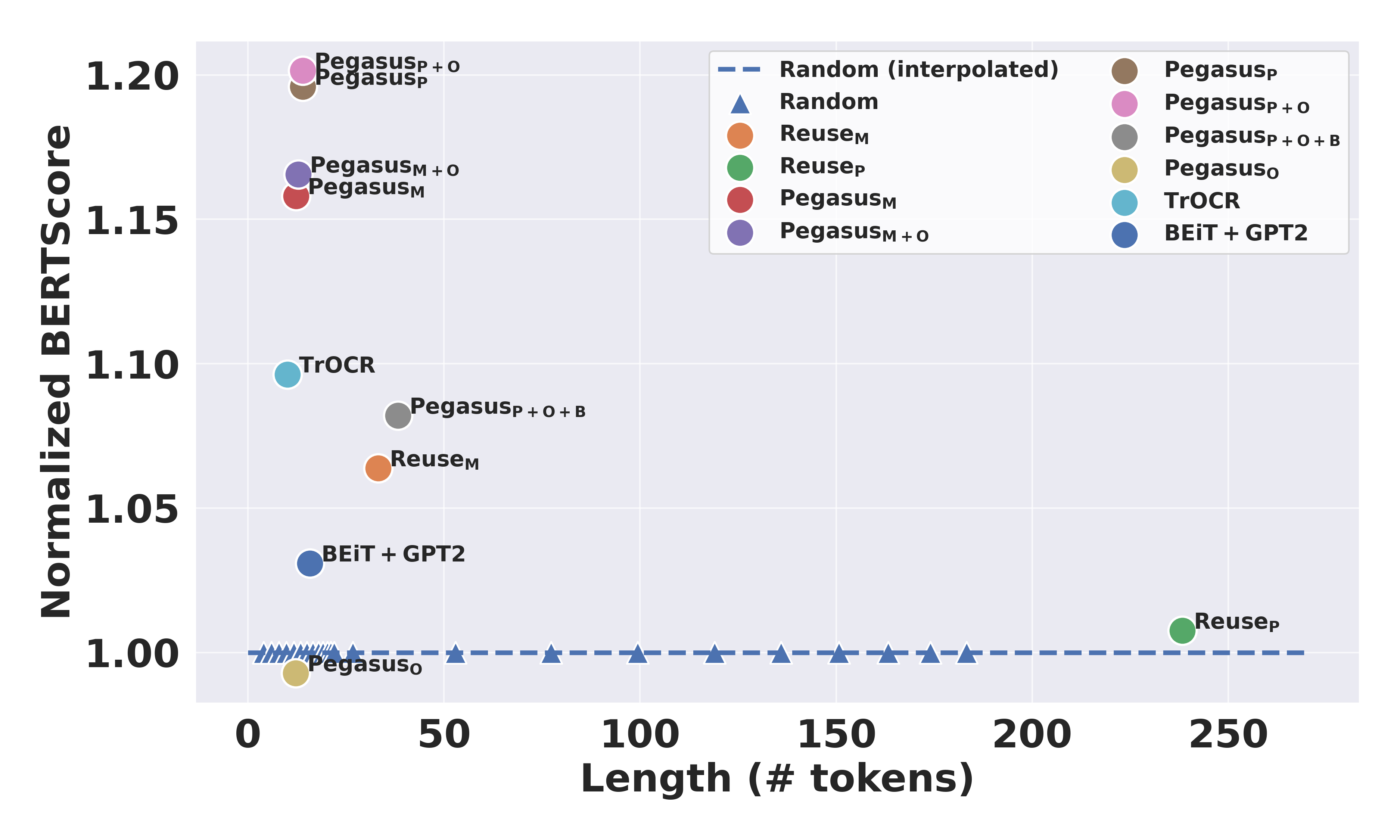}
    \vspace{-3mm}
    \caption{The relationship between average text length and BERTScore. When the generated text is shorter than 30 tokens, longer texts generally result in a higher BERTScore score.} 
    \label{fig:length-performance-bertscore}
\end{figure*}

\paragraph{Examples.}
\Cref{fig:good-bad-example} shows two samples of generation output. The information generated by Pegasus$_{P+O+B}$ could be helpful (A), but it could also introduce factual errors (B).

\begin{figure*}[t]
    \centering
    \includegraphics[width=0.75\linewidth]{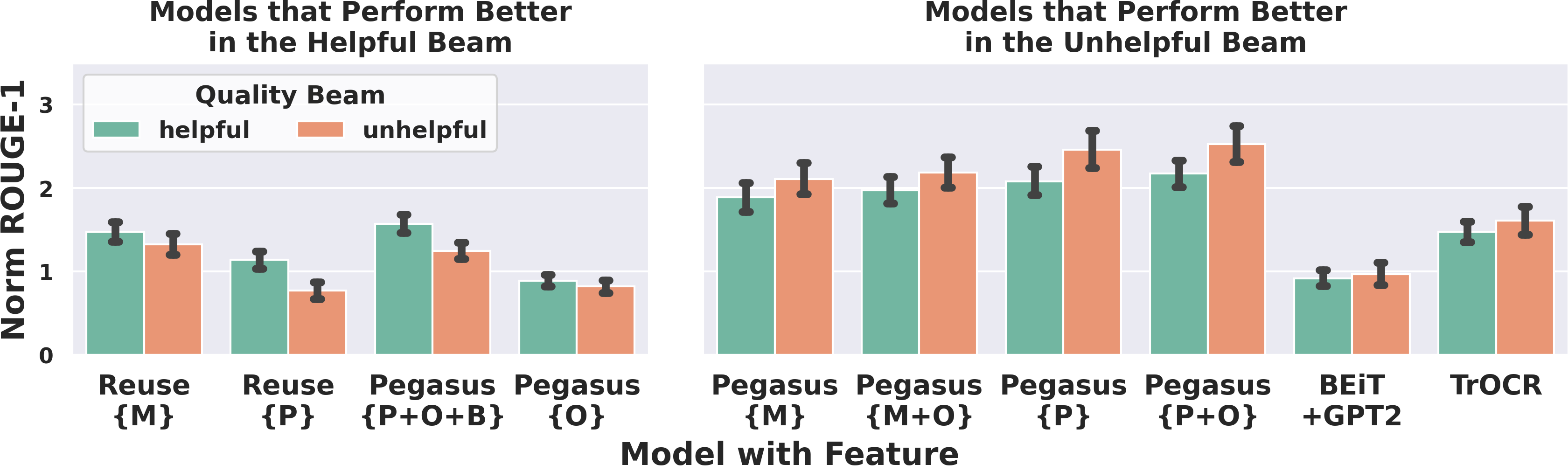} \\
    \vspace{5mm}
    \includegraphics[width=0.75\linewidth]{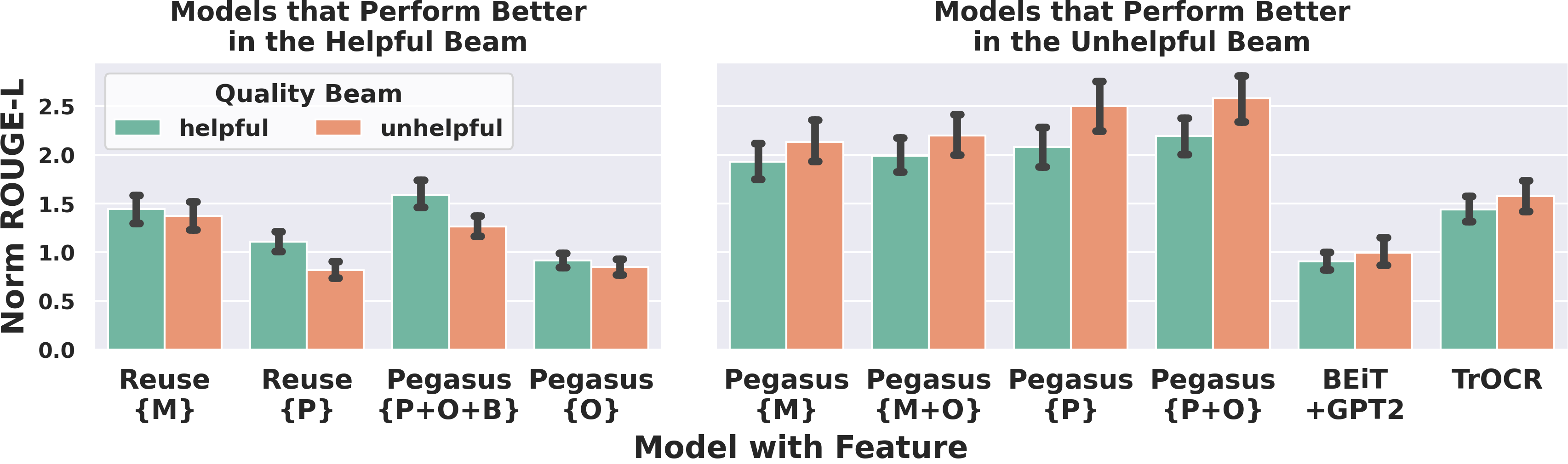} \\
    \vspace{5mm}
    \includegraphics[width=0.75\linewidth]{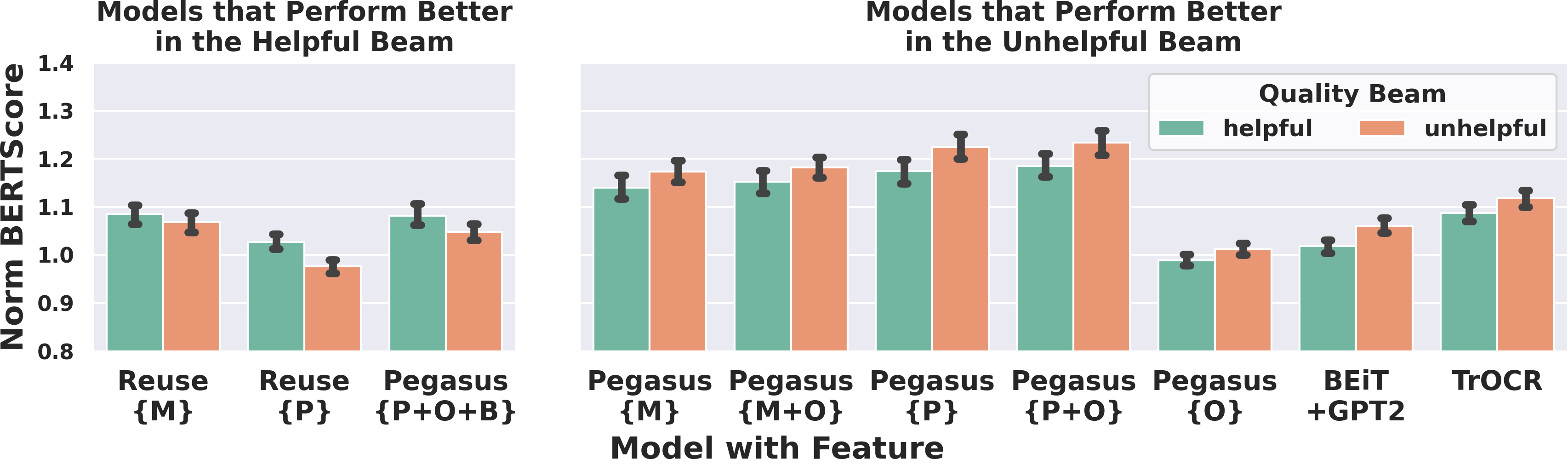}
    \caption{Normalized ROUGE-1, ROUGE-L, and BERTScore for beams of different quality. Most of the generative models (Pegasus, BEiT+GPT2, and TrOCR) performed better in the unhelpful beam, suggesting that they may be better at generating bad captions. Only the model trained with \textbf{better} captions (Pegasus$_{P+O+B}$) learned to generate good captions by showing a much better score in the helpful beam. Note that though Pegasus$_{O}$ also performs better in the helpful beam, the difference is subtle.}
    \label{fig:quality-beam-rouge-rest}
\end{figure*}

\paragraph{Performance in Different Quality Beams.}
\Cref{fig:quality-beam-rouge-rest} shows the ROUGE-1 and ROUGE-L changes in beams of different quality. 
We can see findings similar to \Cref{sec:auto-eval-quality-beam} where among different generation models, only the one trained with data quality control (\ie, Pegasus$_{P+O+B}$) performed better in the helpful beam, generating captions more similar to helpful captions.

\begin{table*}[t]
\centering \small
\begin{tabular}{@{}llcccccc@{}}
\toprule
\multirow{2}{*}{\textbf{Model}} & \multirow{2}{*}{\textbf{Feature}} & \multirow{2}{*}{\textbf{Length}} & \textbf{Rouge-1 (F1)} & \textbf{Rouge-2 (F1)} & \textbf{Rouge-L (F1)} & \textbf{WMS} & \textbf{BERTScore} \\ \cmidrule{4-8}
 &  &  & \textbf{Rand} & \textbf{Rand} & \textbf{Rand} & \textbf{Rand} & \textbf{Rand} \\ \midrule
\multirow{6}{*}{\textbf{Reuse}} & \textbf{M} & 33.2 & .216 & .077 & .171 & .524 & .590 \\
 & \textbf{P} & 238.3 & .164 & .088 & .130 & .501 & .563 \\
 & \textbf{W{[}0, 1{]}} & 50.3 & .231 & .087 & .176 & .520 & .592 \\
 & \textbf{W{[}0, 2{]}} & 68.0 & .230 & .092 & .173 & .518 & .591 \\
 & \textbf{W{[}1, 1{]}} & 67.8 & .230 & .092 & .173 & .518 & .591 \\
 & \textbf{W{[}2, 2{]}} & 98.7 & .217 & .095 & .162 & .513 & .588 \\ \midrule
\multirow{6}{*}{\textbf{Pegasus}} & \textbf{M} & 12.2 & .169 & .053 & .144 & .519 & .565 \\
 & \textbf{M+O} & 12.8 & .173 & .055 & .147 & .520 & .567 \\
 & \textbf{P} & 14.0 & .181 & .058 & .152 & .521 & .570 \\
 & \textbf{P+O} & 14.0 & .181 & .058 & .152 & .521 & .570 \\
 & \textbf{P+O+B} & 38.3 & .221 & .080 & .172 & .523 & .590 \\
 & \textbf{O} & 12.1 & .168 & .052 & .143 & .519 & .565 \\ \midrule
\textbf{TrOCR} & \multirow{2}{*}{\textbf{Figure}} & 10.0 & .150 & .044 & .130 & .517 & .557 \\
\textbf{BEiT+GPT2} &  & 15.8 & .190 & .062 & .158 & .522 & .574 \\
\bottomrule
\end{tabular}
\caption{Random scores corresponding to the length for each automatic evaluation metric.}
\label{tab:random-scores}
\end{table*}

\end{document}